\documentclass[10pt,twocolumn,letterpaper]{article}

\usepackage{cvpr}              

\usepackage{graphicx}
\usepackage{amsmath}
\usepackage{amssymb}
\usepackage{booktabs}
\usepackage{algorithm,algcompatible}
\usepackage[accsupp]{axessibility}  

\usepackage{textcomp, gensymb}
\usepackage{multirow}
\usepackage{amsfonts}
\usepackage[table]{xcolor}
\usepackage{booktabs}
\usepackage{mathtools}
\usepackage{algorithm}
\usepackage{algpseudocode}

\DeclareMathOperator*{\argmin}{arg\,min}
\algnewcommand\INPUT{\item[\textbf{Input:}]}%
\algnewcommand\OUTPUT{\item[\textbf{Output:}]}%

\usepackage[pagebackref,breaklinks,colorlinks]{hyperref}

\usepackage[capitalize]{cleveref}
\crefname{section}{Sec.}{Secs.}
\Crefname{section}{Section}{Sections}
\Crefname{table}{Table}{Tables}
\crefname{table}{Tab.}{Tabs.}

\usepackage{color}

\def\cvprPaperID{7525} 
\def\confName{CVPR}
\def\confYear{2022}

\begin{document}

\title{Robust and Accurate Superquadric Recovery: a Probabilistic Approach}

\author{Weixiao Liu$^{1, 2}$ \quad Yuwei Wu$^{1}$ \quad Sipu Ruan$^{1}$ \quad Gregory S. Chirikjian$^{1}$\footnotemark[1]\\
$^1$National University of Singapore \quad $^2$Johns Hopkins University\\
{\tt\small \{mpewxl, yw.wu, ruansp, mpegre\}@nus.edu.sg}
}
\maketitle

\renewcommand{\thefootnote}{\fnsymbol{footnote}}
\footnotetext[1]{Corresponding author}

\begin{abstract}
   Interpreting objects with basic geometric primitives has long been studied in computer vision. Among geometric primitives, superquadrics are well known for their ability to represent a wide range of shapes with few parameters. However, as the first and foremost step, recovering superquadrics accurately and robustly from 3D data still remains challenging. The existing methods are subject to local optima and sensitive to noise and outliers in real-world scenarios, resulting in frequent failure in capturing geometric shapes. In this paper, we propose the first probabilistic method to recover superquadrics from point clouds. Our method builds a Gaussian-uniform mixture model (GUM) on the parametric surface of a superquadric, which explicitly models the generation of outliers and noise. The superquadric recovery is formulated as a Maximum Likelihood Estimation (MLE) problem. We propose an algorithm, \textit{Expectation, Maximization, and Switching} (EMS), to solve this problem, where: (1) outliers are predicted from the posterior perspective; (2) the superquadric parameter is optimized by the trust-region reflective algorithm; and (3) local optima are avoided by globally searching and switching among parameters encoding similar superquadrics. We show that our method can be extended to the multi-superquadrics recovery for complex objects. The proposed method outperforms the state-of-the-art in terms of accuracy, efficiency, and robustness on both synthetic and real-world datasets. The code is at \url{http://github.com/bmlklwx/EMS-superquadric_fitting.git}.
\end{abstract}

\section{Introduction}
\label{sec:intro}

Understanding 3D environments (and the objects therein) has been a crucial task for computer vision and intelligent systems\cite{pentland1987perceptual}.
With the recent breakthroughs in 3D vision and the advancement of computing power, modern computer vision systems are able to reconstruct and reason about scenes with low-level representations, such as point clouds\cite{fan2017a, qi2017pointnet, qi2017pointnetplusplus}, meshes\cite{lorensen1987marching} and voxels\cite{voxel96, zeng20163dmatch}.
In contrast, the human visual system favors abstracting scenes into canonical parts for a better perceptual understanding\cite{core_systems_in_human_cognition, biederman1987recognition}.
Human vision seems to work well with basic geometric structures of the scene, and not rely on detailed point-wise models\cite{pentland1987perceptual}.
Therefore, understanding the geometric structure of visual inputs appears to be a promising approach for intelligent systems to achieve high-level goals such as physical reasoning, decision making, planning and interaction with the environment.
Inspired by this idea, researchers turned to exploring the possibility of using basic volumetric primitives for object description, \eg, cuboids\cite{Tulsiani_2017_CVPR, cube_cvpr2017, cube_cvpr2018}.
However, due to the limited expressiveness of cuboids, objects can only be describe in a highly abstracted way.
\begin{figure} [!tp]
    \centering
    \includegraphics[width=0.80\columnwidth]{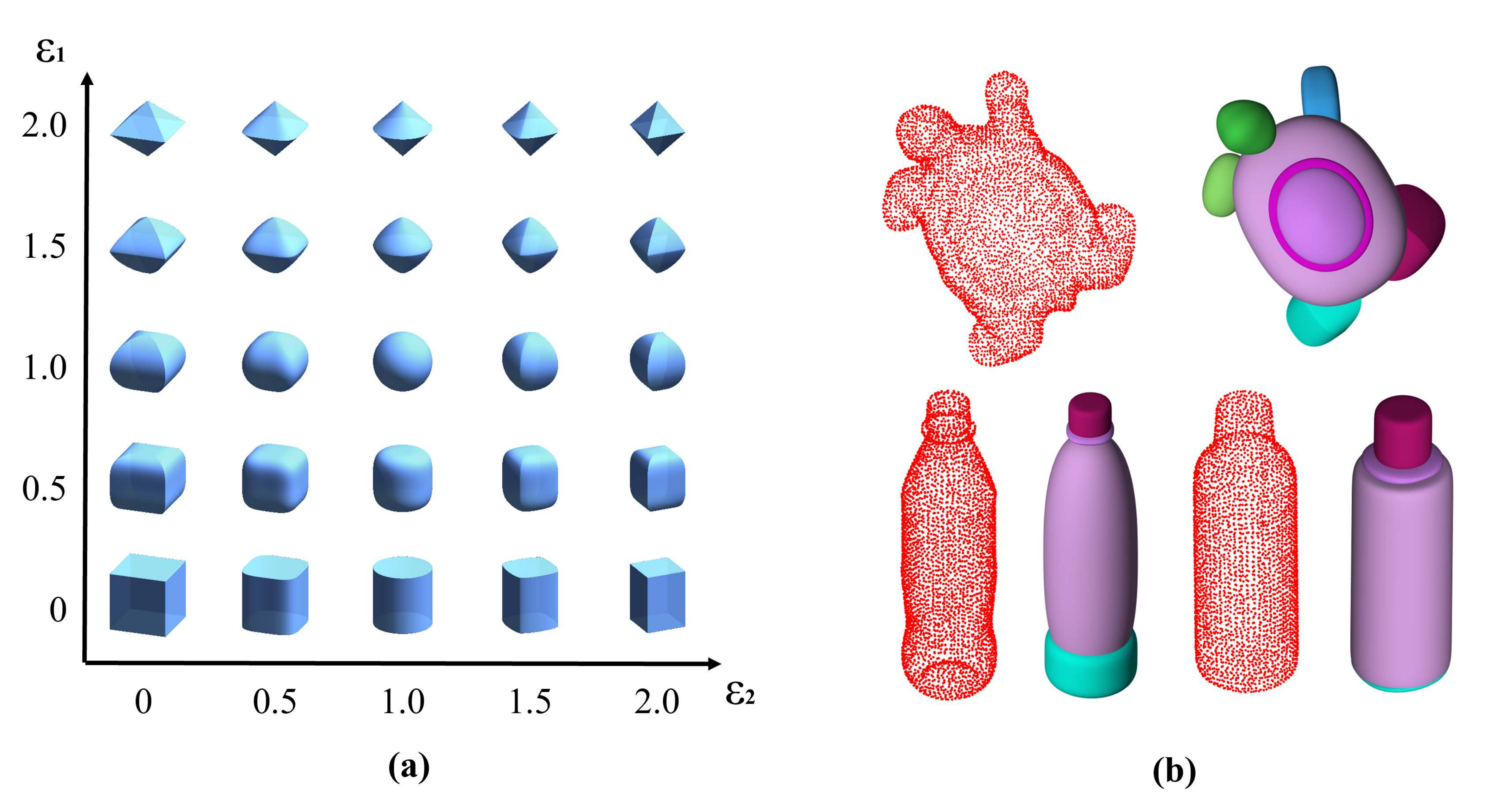} 
    \caption{(a) Shape vocabulary of convex superquadrics. We focus on the interpretation of point clouds with convex superquadrics. (b) Superquadric representations inferred by our method.}
    \label{fig:superquadrics}
    \vspace{-0.5cm}
\end{figure}

Superquadrics are a family of geometric primitives with a rich shape vocabulary, including cuboids, cylinders, ellipsoids, octahedra and their intermediates (Fig. \ref{fig:superquadrics}), but only encoded by 5 parameters.
In the recent few years, superquadrics have raised considerable attention in the community and are widely applied in robotics and computer vision tasks, \eg, object modeling\cite{paschalidou2019superquadrics, paschalidou2020learning, BORGES1997655}, collision detection and motion planning \cite{ruan2019efficient, RUAN_closedform}, pose estimation \cite{ObjectDetectionICRA}, and grasping \cite{vezzani2017grasping, vezzani2018improving, quispe2015exploiting}.
A single superquadric is already expressive enough to reasonably model many everyday objects\cite{kasper2012kit, singh2014bigbird}.
The pioneering work on single superquadric recovery includes \cite{Bajcsy1987ICCV, solina1990recovery, GrossICCV88}, where the recovery is formulated as a \textit{Least-Square} (LSQ) problem.
In \cite{leonardis1997superquadrics, chevalier2003segmentation}, the methods are extended to model complex objects with multiple superquadrics.
However, the existing methods are vulnerable to noise and outliers, and the inherent deficiency of the local optimizer makes them sensitive to initialization.

In this paper, we focus on the single superquadric recovery problem and propose the first probabilistic method to recover superquadrics from noisy point clouds.
We build a probabilistic model to simulate how observations (point cloud) are possibly generated from a given parameterized superquadric surface.
GUM is adopted to accommodate the generation of noise and outliers.
The superquadric recovery is thus formulated as a MLE problem (Sec. \ref{sec:probabilistic_formulation}).
\textit{Expectation and Maximization} (EM) algorithm is most commonly used to solve MLE problems with latent variables\cite{DEMP1977}.
However, the EM algorithm typically converge to a local optimum, resulting in an inaccurate shape approximation.
To solve this problem, we propose a novel algorithm, \textit{Expectation, Maximization and Switch} (EMS), which takes advantage of the geometric features of superquadrics to avoid local optima (Sec. \ref{sec:ems}).
In the E-step, the probability of a point being an outlier is inferred from a posterior perspective (Sec. \ref{sec:expectation}).
In the M-step, the parameter of the superquadric is updated given the current estimation of the latent variables in the E-step (Sec. \ref{sec:maximization}).
The E and M-step alternate until converged to a local optimum, where we introduce the S-step: globally search for candidate parameters encoding similar superquadrics in terms of shape and pose, and then switch to the one from which we can further increase the likelihood (Sec. \ref{sec:switch}).
This strategy is made possible by investigating the geometric symmetry and algebraic ambiguity of superquadrics.
Furthermore, we show our probabilistic formulation can be extended to multi-superquadrics recovery tasks (Sec. \ref{sec:extension}).

The motivations and key features of this paper come from the two following questions.
(1) Can we design an algorithm robust, accurate and efficient enough to abstract an optimal superquadric primitive from a noisy point cloud whose underlying shape is indeed within the vocabulary of superquadrics?
This question is answered in Sec. \ref{experiment_synthetic} and Sec. \ref{experiment_real}.
(2) If the point cloud is too complex to be approximate by a single superquadric, can the algorithm figure out a major part which can be approximated by a superquadric?
This is a challenging goal which cannot be achieved by any existing single superquadric recovery algorithm.
We demonstrate qualitatively that the proposed method is capable of doing this in Sec. \ref{sec:multi-superquadrics}.

\section{Related work}
\label{sec:related_wrok}
\subsection{Preliminary: Superquadrics}
\label{preliminary}
Superquadrics are a family of geometric primitives (Fig. \ref{fig:superquadrics}), which can be defined by the implicit function\cite{barr1981superquadrics}:
\begin{equation}
    F(\mathbf{x})\doteq\left(\left(\frac{x}{a_{x}}\right)^{\frac{2}{\epsilon_{2}}}+\left(\frac{ y}{a_{y}}\right)^{\frac{2}{\varepsilon_{2}}}\right)^{\frac{\varepsilon_{2}}{\varepsilon_{1}}}+\left(\frac{z}{a_{z}}\right)^{\frac{2}{\varepsilon_{1}}}
    \label{eqn_iofunction}
\end{equation}
where $\mathbf{x}\doteq[x, y, z]^T\in \mathbb{R}^3$ is a point defined in the superquadric frame;
$a_x$, $a_y$ and $a_z\in \mathbb{R}_{>0}$ are the scale parameters corresponding to the x, y, z-axis;
$\epsilon_1$ and $\epsilon_2\in \mathbb{R}_{\geq0}$ are the shape parameters.
$\epsilon_1$ governs the shape along the z-axis and $\epsilon_2$ controls the shape orthogonal to the z-axis.
A superquadric is convex, if both $\epsilon_1$ and $\epsilon_2$ are within $(0, 2]$.
Eq. \eqref{eqn_iofunction} is also called the `inside-outside' function, since $\mathbf{x}$ locates on the surface if $F(\mathbf{x})=1$, inside if $F(\mathbf{x})<1$ and outside otherwise.
We are able to fully parameterize a superquadric of a general pose with $\boldsymbol{\theta}\doteq \{\epsilon_1, \epsilon_2, a_x, a_y, a_z, g\}$, where $g=[\mathbf{R}\in SO(3), \mathbf{t}\in \mathbb{R}^3]\in SE(3)$ is the Euclidean transformation.

\subsection{Superquadric Recovery}
The single superquadric recovery plays an important role in all superquadric recovery problems.
In the previous works\cite{Bajcsy1987ICCV, solina1990recovery, GrossICCV88, superquadrics_num_stable, hu1995robust}, it is formulated as a LSQ problem and solved with the Levenberg-Marquardt (LM) algorithm\cite{levenberg1944method}.
Recently, researchers have also explored the possibility of deep learning approaches\cite{Oblak_access, sirceljsegmentation}.
Solina \etal. \cite{Bajcsy1987ICCV,  solina1990recovery} propose a cost function based on the implicit function of superquadrics.
Gross \etal. \cite{GrossICCV88} modifies the cost function with the radial distance between points and the superquadric surface.
In \cite{vanDop98CVPR, hu1995robust}, the robustness of the algorithms is investigated, and outliers are randomly rejected by a customized heuristic function.
All the above methods suffer from numerical instability when either $\epsilon_1$ or $\epsilon_2$ approaches 0.
As a consequence, they compromise by constraining the lower bounds of $\epsilon_1$ and $\epsilon_2$ to 0.1, resulting in less accuracy when modeling shapes with sharp edges, \textit{e.g.}, cuboids and cylinders.
Recently, this problem is revisited and solved by approximating the implicit function with an auxiliary function in the unstable region\cite{superquadrics_num_stable}.
Also in \cite{superquadrics_num_stable}, the authors point out that the methods are sensitive to the initialization, and suggest to try 3 different initial guesses obtained by performing the Principal Component Analysis (PCA) to the point cloud.
However, all the methods are still subject to local optima, especially when the point cloud is partially sampled on the object surface.

For the multi-superquadrics recovery, Leonardis \cite{leonardis1997superquadrics} Borges \cite{BORGES1997655}, and Chevalier \etal.\cite{chevalier2003segmentation} try to interpret complex objects with multiple superquadrics by first segmenting the point cloud into parts and fitting a superquadric to each of them.
In contrast, our multi-superquadrics recovery extension works in a hierarchical way.
Other than a multi-superquadrics representation, we can also obtain the hierarchical relationships among the superquadrics (Fig. \ref{fig:hierarchical}).

Another line of the multi-superquadrics recovery is based on deep learning \cite{paschalidou2019superquadrics, paschalidou2020learning}, where superquadrics can be obtained from meshes or even an RGB image.
However, their approaches are data-driven with limitations in generalization, while we focus on the recovery of \textit{unseen} real-world objects based on a case by case probabilistic reasoning.

\section{Our Method}
\subsection{Probabilistic Formulation}
\label{sec:probabilistic_formulation}
In this section, we demonstrate how to formulate the superquadric recovery as a MLE problem.
Our model is built as follows.
First, on the superquadric surface $\mathcal{S}_{\boldsymbol{\theta}}\subset \mathbb{R}^3$ parameterized by $\boldsymbol{\theta}$, a Gaussian centroid $\boldsymbol{\mu}\in \mathcal{S}_{\boldsymbol{\theta}}$ is sampled randomly according to the uniform density function
\begin{equation}
    p(\boldsymbol{\mu}) = \frac{1}{{A}_{\boldsymbol{\theta}}}, \quad A_{\boldsymbol{\theta}}=\int_{\mathcal{S}_{\boldsymbol{\theta}}}1dS
\end{equation}
where $A_{\boldsymbol{\theta}}$ is the area of the superquadric surface.
Subsequently, an observation $\mathbf{x}\in\mathbb{R}^3$ is generated from a Gaussian-uniform model (GUM), the probability density function of which is given as
\begin{equation}
    p(\mathbf{x}|\boldsymbol{\mu}) = w_op_o(\mathbf{x}) + (1-w_o)\mathcal{N}(\mathbf{x}|\boldsymbol{\mu}, \mathbf{\Sigma})
    \label{eqn_gum}
\end{equation}
where $\mathcal{N}(\cdot|\boldsymbol{\mu}, \mathbf{\Sigma})$ represents the density function of a Gaussian distribution with mean $\boldsymbol{\mu}$ and covariance $\mathbf{\Sigma}$.
We assume the noise to be isotropic, \textit{i.e.}, $\boldsymbol{\Sigma} = \sigma^2\mathbf{I}$, where $\mathbf{I}\in \mathbb{R}_{3\times 3}$ is the identity matrix.
A uniform outlier component is introduced to model the generation of outliers.
$w_o\in[0, 1]$ is the probability of a point being sampled from the outlier component.
To make Eq. \eqref{eqn_gum} a proper probability density function (integrates to 1 over $\mathbb{R}^3$), we introduce a working space $\mathbb{V}$ encapsulating the point set, whose volume equals $V$.
The probability density of the outlier component $p_o(\mathbf{x}) = \frac{1}{V}$ if $\mathbf{x}\in \mathbb{V}$ and $p_o(\mathbf{x}) = 0$ otherwise.

For the convenience of the upcoming derivation, we reformulate Eq. \eqref{eqn_gum} into an equivalent form by introducing a latent random variable $z\in\{0,1\}$ encoding the membership of $\mathbf{x}$.
When $z=0$, $\mathbf{x}$ is sampled from the uniform outlier component. On the contrary, $z=1$ indicates that $\mathbf{x}$ is generated from the Gaussian inlier component.
We assume the membership $z$ is independent of the mean $\mathbf{\mu}$ of the Gaussian component.
Therefore, Eq. \eqref{eqn_gum} is equivalent to
\begin{equation}
\begin{gathered}
     p(\mathbf{x}|\boldsymbol{\mu}, z) = p_o(\mathbf{x})^{1-z}\cdot\mathcal{N}(\mathbf{x}|\boldsymbol{\mu}, \mathbf{\Sigma})^{z}\\
     z\sim p(z)=\mathrm {Bernoulli}(1-w_o)
\end{gathered}
\label{eqn_gum_latent}
\end{equation}

Given a set of points $\mathbf{X} = \{\mathbf{x}_i\in \mathbb{R}^{3}|i=1,2,...,N\}$, the parameters of the superquadric surface can be estimated by maximizing the following likelihood function
\begin{equation}
    L(\boldsymbol{\theta}, \sigma^2)=\prod_{i = 1}^{N}p(\mathbf{x}_i|\boldsymbol{\mu}_i,z_i)p(\boldsymbol{\mu}_i)p(z_i)
\end{equation}
or equivalently, minimizing the negative log-likelihood function
\begin{equation}
    l(\boldsymbol{\theta}, \sigma^2)=\sum_{i = 1}^{N}z_i \, \left( \frac{\|\mathbf{x}_i-\boldsymbol{\mu}_i\|_2^2}{2\sigma^2}-\log c \right)+N\log(A_{\boldsymbol{\theta}})
    \label{eqn_loglikelihood}
\end{equation}
where $c$ is the normalizing constant of the Gaussian distribution defined in Eq. \eqref{eqn_gum}.
Note that the terms independent of $\boldsymbol{\theta}$ and $\sigma^2$ have been omitted for simplicity.

\subsection{Expectation, Maximization and Switching}
\label{sec:ems}
Eq. \eqref{eqn_loglikelihood} is intractable to solve directly because of the existence of a set of continuous latent  variables $\boldsymbol{\mu}_i\in \mathcal{S}_{\boldsymbol{\theta}},i=1,2,...,N$, and discrete latent variables $z_i\in\{0,1\}, i=1,2,...,N$.
The EM algorithm\cite{DEMP1977} is most commonly used to solve MLE with latent variables.
However, in our scenario, it suffers from multiple local optima and has difficulties in calculating the expectation of the latent variables in an exact form.
Therefore, we propose a novel algorithm called EMS, which is adapted from the EM algorithm by utilizing the geometric properties of superquadrics. 
The algorithm first estimates the latent variables from a posterior perspective given the current estimation of the superquadric parameter (E-step).
Next, the parameters are inferred and updated by minimizing the negative log-likelihood function (M-step).
These two steps alternate until converged to a local minimum.
Then, we globally search for candidate parameters which encode superquadrics geometrically similar to the current estimation but have the potential of further decreasing the negative log-likelihood function (S-step).
If a valid candidate is found, we reinitialize the EM-phase with this candidate;
else, we declare a termination and return the final result.
Detailed motivations and derivations of each step are discussed in the following sections.

\subsection{E-step: Latent Variables Estimation}
\label{sec:expectation}
Given the current superquadric parameter $\boldsymbol{\theta}$, we estimate $\boldsymbol{\mu}_i$ via \textit{maximum a posteriori} (MAP)
\begin{equation}
    \hat{\boldsymbol{\mu}}_i=\argmin_{\boldsymbol{\mu}_i\in \mathcal{S}_{\boldsymbol{\theta}}}p(\boldsymbol{\mu}_i|\mathbf{x}_i)
\end{equation}
which guarantees to minimize Eq. \eqref{eqn_loglikelihood}.
By the Bayes' rule, we show that it is equivalent to find the nearest point on the superquadric surface to $\mathbf{x}_i$
\begin{equation}
    \hat{\boldsymbol{\mu}}_i = \argmin_{\boldsymbol{\mu}_i\in \mathcal{S}_{\boldsymbol{\theta}}}\|\mathbf{x}_i-\boldsymbol{\mu}_i\|_2
    \label{eqn_closestpoint}
\end{equation}
As far as we know, there is no analytical solution to Eq. \eqref{eqn_closestpoint}, and numerical optimization is prohibitively time consuming for a large number of observations.
Therefore, we approximate $\hat{\boldsymbol{\mu}}_i$ with a sub-optimal solution
\begin{equation}
    \hat{\boldsymbol{\mu}}_i^s=g\circ\left[F^{-\frac{\epsilon_1}{2}}(g^{-1}\circ\mathbf{x}_i) \cdot g^{-1}\circ\mathbf{x}_i\right]
    \label{eqn_radial_point}
\end{equation}
which is the intersection of the vector $\mathbf{x}_i$ (defined in the superquadric frame) and the superquadric surface $\mathcal{S}_{\boldsymbol{\theta}}$.
$F(\cdot)$ is the implicit function of the superquadric.
This idea was first introduced by \cite{GrossICCV88} to establish the radial distance measure for superquadrics fitting.

Given the current estimation of $\boldsymbol{\mu}_i$, the expectation of the posterior probability of $\mathbf{x}_i$ being an inlier is inferred via the Bayes' rule 
\begin{equation}
    E(z_i=1|\mathbf{x}_i, \hat{\boldsymbol{\mu}}_i^s)=
    \frac{\mathcal{N}(\mathbf{x}_i|\hat{\boldsymbol{\mu}}_i^s, \sigma^2\mathbf{I})}{\mathcal{N}(\mathbf{x}_i|\hat{\boldsymbol{\mu}}_i^s, \sigma^2\mathbf{I}) + \frac{w_op_o}{1-w_o}}
    \label{eqn_expectation_z}
\end{equation}
For simplicity, we denote the value of the expectation as $\hat{z}_i$.

\subsection{M-step: Parameter Optimization}
\label{sec:maximization}
Substituting $\boldsymbol{\mu}_i$ and $z_i$ in Eq. \eqref{eqn_loglikelihood} with the their posterior estimations $\hat{\boldsymbol{\mu}}_i^s$ and $\hat{z}_i$, we obtain an updated negative log-likelihood function
\begin{equation}
    \hat{l}(\boldsymbol{\theta}, \sigma^2)=\sum_{i = 1}^{N}\hat{z}_i \, \left(\frac{\|\mathbf{x}_i-\hat{\boldsymbol{\mu}}_i^s\|_2^2}{2\sigma^2}-\log c \right)+N\log(A_{\boldsymbol{\theta}})
    \label{eqn_loglikelihood_update}
\end{equation}
To the best of our knowledge, $A_\theta$ cannot be expressed in closed-form with respect to $\boldsymbol{\theta}$.
For efficiency, $A_\theta$ is approximated by a bi-linear interpolation along $\epsilon_1$ and $\epsilon_2$.
Details are presented in the Supplementary Material. 
$N\log(A_{\boldsymbol{\theta}})$ indicates a preference for a small superquadric.
To avoid local optima, it is preferable to delay the introduction of this term until $\sigma^2$ falls below a threshold, \textit{i.e.}, the shape is roughly captured.
We apply the trust-region-reflective \cite{trust_region_reflective} to optimize $\boldsymbol{\theta}$.
The main reason we do not use the LM solver lies in its numerical instability when approaching the lower bound of the shape parameters.
In contrast, the reflective transformation\cite{reflective94}, empirically, always keeps the optimization numerically stable.
More discussion can be found in the Supplementary Material.
After $\boldsymbol{\theta}$ is solved, $\sigma^2$ is updated by setting the the corresponding partial derivative of Eq. \eqref{eqn_loglikelihood_update} to zero.
\subsection{S-step: Geometric Local Optimum Avoidance}
\label{sec:switch}
\begin{figure} [!ht]
    \centering
    \includegraphics[width=0.9\columnwidth]{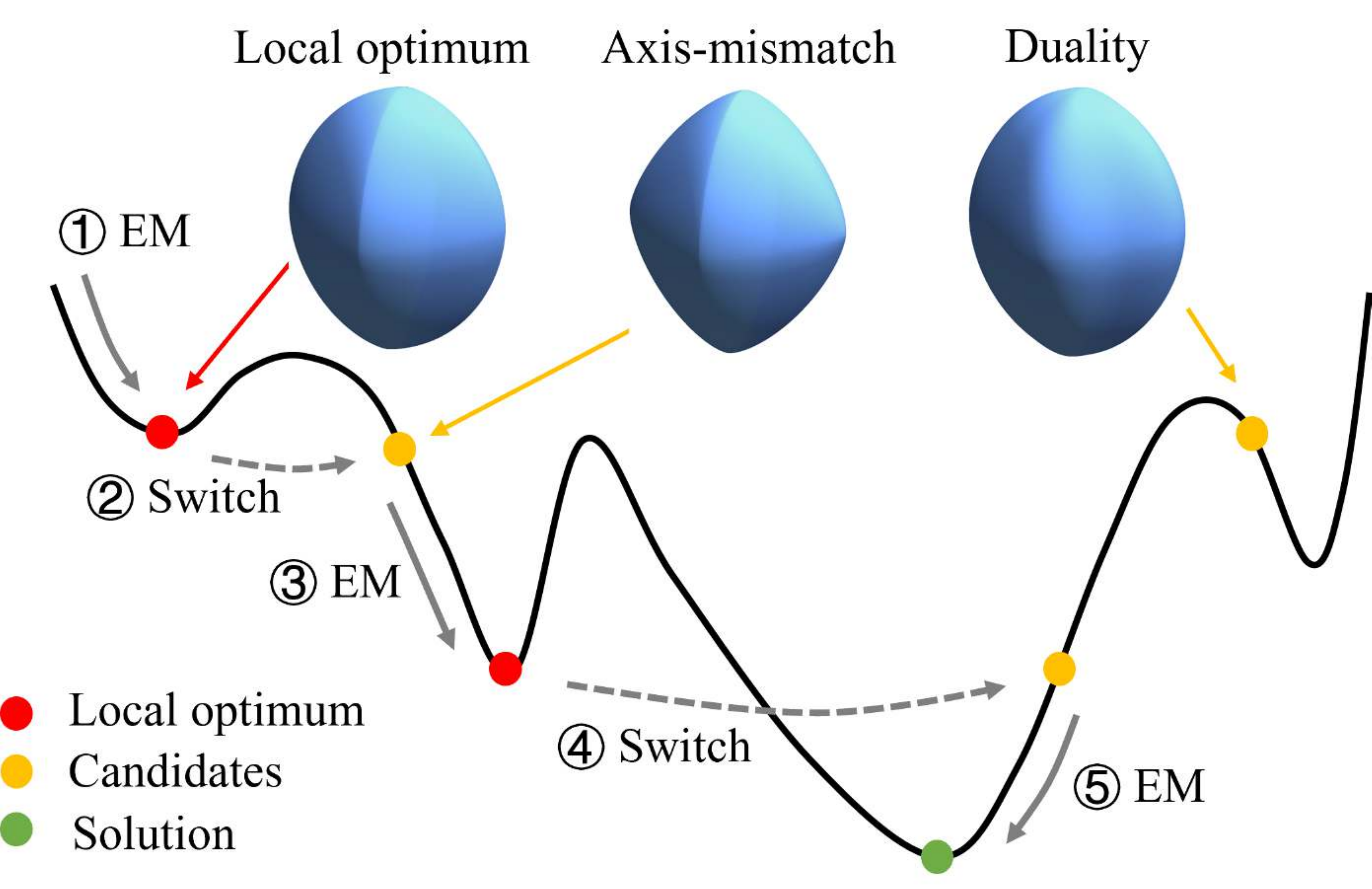} 
    \caption{Visualization of the S-step and similarities. The EM steps change the shape and pose of the superquadric towards a local optimum, while the S-steps tries to preserve the current shape but jump between distant parameters to avoid the local optimum.}
    \label{fig:s_step}
    \vspace{-0.5cm}
\end{figure}

Superquadrics hold many geometric and algebraic features, \textit{e.g.}, symmetry and parametric ambiguity.
Trivial cases can be raised where two superquadrics possess a similar or even identical shape and pose but are distant in the parameter space.
The likelihood function Eq. \eqref{eqn_loglikelihood_update} quantifies how well an underlying geometric primitive fits the observations, and thus two similar (or identical) superquadrics must possess similar (or identical) likelihoods as well.
Inspired by this idea, we propose a geometry-guided local optima avoidance strategy: globally search for candidate parameters encoding similar shapes, and switch to the one which has the potential of further increasing the likelihood.
The switch action is discontinuous in the parameter space, while the change in terms of the underlying geometric shape is almost smooth.
We investigate the conditions when two parameters distant to each other represent similar superquadrics, and summarize the cases into two categories.
The idea of switch and examples of similarities are visualized in Fig. \ref{fig:s_step}.

\textbf{Axis-mismatch similarity}: According to the definition in Sec. \ref{preliminary}, the x and y-axis are of an equivalent class in terms for interchangeability, \textit{i.e.}, if we interchange the value of $a_x$ and $a_y$, we can obtain an identical superquadric by rotating 90 degrees about the z-axis.
However, the z-axis distinguishes itself because it is generally not interchangeable with them.
We thus name the z-axis as the principal axis.
Now let us consider a special case when the shape parameters $\epsilon_1 = \epsilon_2$.
In this scenario, the implicit function (Eq. \eqref{eqn_iofunction}) degenerates and makes all the three axes equivalent.
From another perspective, the `degree' of interchangeability decrease as the difference between $\epsilon_1$ and $\epsilon_2$ increases.
Therefore, when $\epsilon_1$ and $\epsilon_2$ are close to each other, a similar superquadric can be obtained by reassigning the principal axis to either the x or y-axis and then applying a corresponding rotation.
We call this axis-mismatch, because it helps to avoid the local optimum caused by the incorrect assignment of the principal axis, which prohibit the shape from further varying to the global optimum.

\textbf{Duality similarity}: This case is derived from the shape ambiguity regarding $\epsilon_2$.
When $a_x=a_y$ and $\epsilon_2=0$, we can always represent the same superquadric (in terms of shape and pose) by scaling $a_x$ and $a_y$ with $1/\sqrt{2}$, setting $\epsilon_2=2$, and rotating 45 degrees about the principal axis.
This is because $\epsilon_2=0$ and $\epsilon_2=2$ define a pair of squares dual to each other in the subspace orthogonal to the principal axis.
Utilizing this property, when $a_x\approx a_y$, a similar superquadric can be constructed by setting $\epsilon_2=2-\epsilon_2$, re-scaling and rotating.
The duality similarity is of great importance to local optimum avoidance, since it connects two extreme ends of the parameter space.
Local optimizer tends to converge monotonically towards one of the ends, while the duality similarities glue the ends together, allowing the algorithm to check and explore the counterpart.

Similar superquadrics can also be resulted by a combination of the two similarities.
Detailed mathematical formulation about the candidates generation and switching strategy can be found in the Supplementary Material.

\subsection{Extension: Multi-superquadrics Recovery}
\label{sec:extension}
\begin{figure}[!ht]
    \centering 
    \includegraphics[width=0.8\columnwidth]{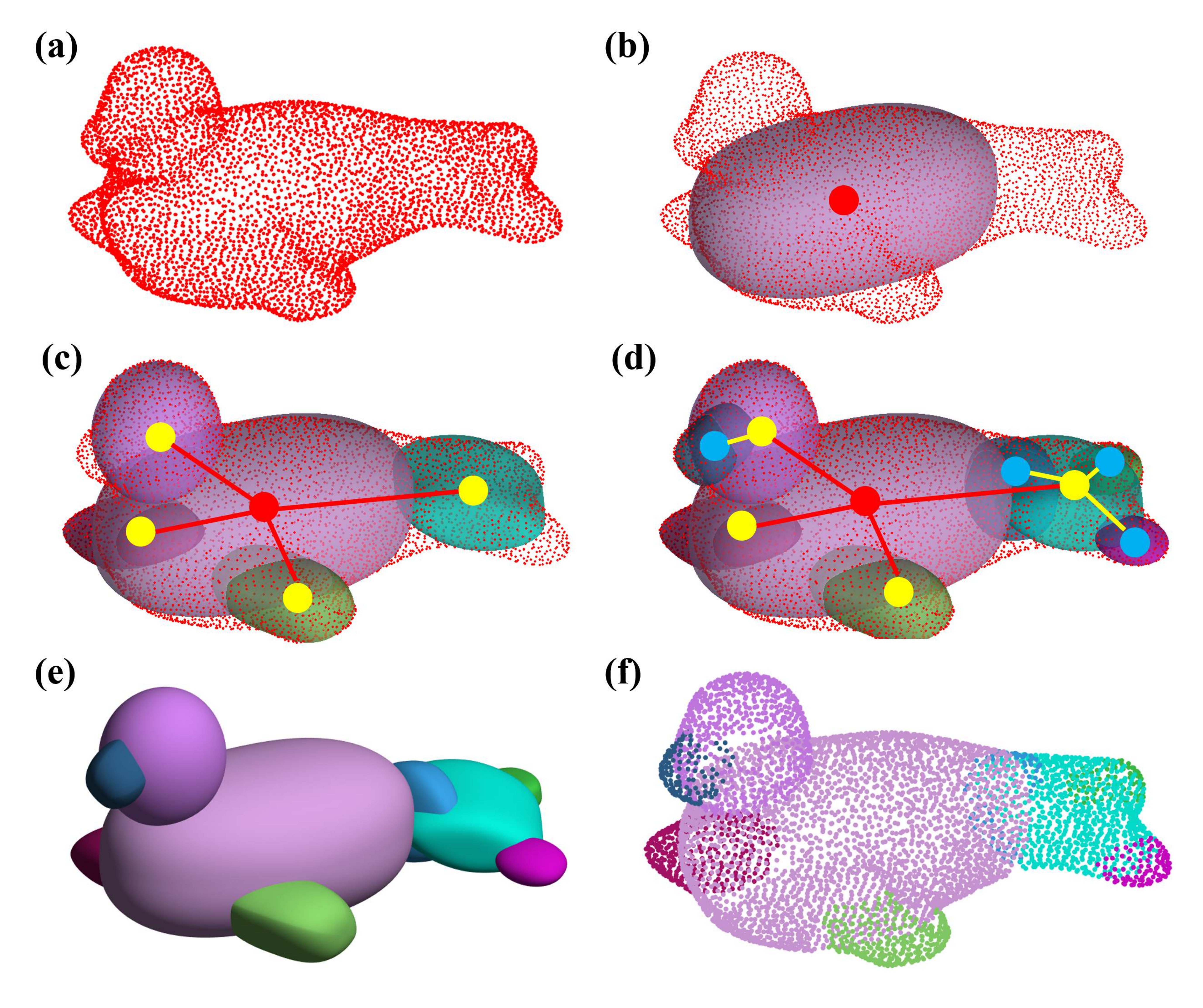} %
    \caption{Hierarchical process of the multi-superquadrics recovery. We demonstrate with a \textit{Seal} from the KIT ObjectModels. (a) The original point cloud. (b-d) Steps of the hierarchical process and the generation of the structure graph. (e) The recovered superquadric model. (f) Segmentation of the original point cloud based on the superquadric model.}
    \label{fig:hierarchical}
    \vspace{-0.4cm}
\end{figure}

Although our method is designed for single superquadric recovery, it can be easily extended to understand the shape and structure of a complex object with multiple superquadrics.
One of the key features of our single superquadric recovery method is that it can capture the major superquadric-like part of a point cloud and identify the unfitted points as `outliers'.
Therefore, from a point cloud $\boldsymbol{X}$ of a complex object, we can first recover a superquadric capturing the major superquadric structure along with a set of unfitted `outlier' clusters $\mathcal{C}$. 
Then, we feed $\mathcal{C}$ back to the algorithm as the input to recover a new set of superquadrics along with an updated $\mathcal{C}$. 
The process (Algorithm \ref{alg:msq}) is repeated hierarchically until $\mathcal{C}$ is empty or a maximum depth of layer is reached.
In this way, not only the point cloud is represented with the union of the acquired superquadrics, but also the hierarchical relationship between the superquadrics is obtained, as illustrated in Fig. \ref{fig:hierarchical}.
\begin{algorithm}
\caption{A layer of hierarchical superquadric recovery}\label{alg:msq}
\begin{algorithmic}[1]
    \State \textbf{Input:} $\{\boldsymbol{X}_1,\boldsymbol{X}_2,...,\boldsymbol{X}_n\}$
    \State \textbf{Output:} $\{ \boldsymbol{\theta}_1,\boldsymbol{\theta}_2,...,\boldsymbol{\theta}_n \}, 
    \mathcal{C} = \{\Bar{\boldsymbol{X}}_1,\Bar{\boldsymbol{X}}_2,...,\Bar{\boldsymbol{X}}_m\}$
    \State $\mathcal{C} \gets \{\}$
    \For{$i=1,...,n$}
        \State $(\boldsymbol{\theta}_i,outliers) \gets EMS(\boldsymbol{X}_i)$
        \State $\{\Bar{\boldsymbol{X}}_{1_i},\Bar{\boldsymbol{X}}_{2_i},...,\Bar{\boldsymbol{X}}_{\gamma_i}\} \gets Clustering(outliers)$ \\
        \Comment{clustering the outliers by Euclidean distance \cite{MatlabOTB}}
        \State $\mathcal{C} \gets \mathcal{C} \cup \{\Bar{\boldsymbol{X}}_{1_i},\Bar{\boldsymbol{X}}_{2_i},...,\Bar{\boldsymbol{X}}_{\gamma_i}\}$\\
        \Comment{clusters with points less then a threshold are pruned}
    \EndFor
\end{algorithmic}
\end{algorithm}

\section{Experiments}
We conduct experiments on both synthetic and real-world point cloud datasets to verify the accuracy, efficiency and robusteness of our method.
All the algorithms and experiments are implemented in MATLAB, on a computer running Intel Core i9-9900K(3.6GHz).
In all the single superquadric recovery experiments, we compare with 4 baseline methods: Implicit-LSQ\cite{solina1990recovery}, Radial-LSQ\cite{GrossICCV88}, Robust-fitting\cite{hu1995robust} and Numerical Stable method\cite{superquadrics_num_stable}.

\textbf{Initialization:}
Following \cite{superquadrics_num_stable}, if not specified, the baseline methods are initialized with 3 different superquadric parameters obtained via PCA to alleviate the affect of local optima.
To show the effectiveness of the proposed local optimum avoidance strategy, our method is only initialized once with the first one among the three. 
Implementation details can be found in the Supplementary Material.

\subsection{Recovery on Synthetic Datasets}
\label{experiment_synthetic}
In this section, we evaluate the performance of the algorithms on recovering the underlying geometric primitives given points sampled on a superquadric.
Three experiments are designed to evaluate the robustness to partial data, outliers, and noise, respectively.
To cover the whole parameter space of superquadrics in convex region, we randomly generate 500 superquadrics under the following strategy:
the shape parameter $[\epsilon_1, \epsilon_2]$ is uniformly sampled on $(0, 2]^2\subset \mathbb{R}^2$;
the scale parameter $[a_x, a_y, a_z]$ is uniformly sampled on $[0.5, 3]^3\subset \mathbb{R}^3$;
the rotation $\mathbf{R}\in SO(3)$ is generated by first sampling a rotation axis on the unit sphere and then applying a random rotation in $[0, 2\pi]$ about the axis;
the translation $\mathbf{t}$ is uniformly sampled on $[-1, 1]^3\subset \mathbb{R}^3$.
Point clouds are generated by sampling evenly on the surface of the random superquadrics with an equal\footnote[2]{No strict and theoretical way of equal-distance sampling exists. Here we indicate an practical approximation of almost equal-distance sampling.} interval of 0.2.
In \cite{superquadrics_num_stable}, a recursive algorithm is proposed to realize this task.
However, the points sampled by the algorithm get denser when approaching the poles along the z-axis.
Therefore, we further improve their algorithm by adaptively re-scaling the sampling intervals at different latitudes to achieve an overall equal-distance sampling (see Supplementary Material).

The most natural metric to evaluate the goodness of fitting is the average point-to-surface orthogonal distance.
However, given an arbitrary point, there is no analytical solution of the corresponding closest point on a superquadric surface.
Therefore, we approximate the point-to-surface metric with
\begin{equation}
    error(\mathbf{X}, \mathbf{S}) \doteq \frac{1}{N} \sum_{i=1}^{N} \min_{\mathbf{s}_j\in\mathbf{S}} \|\mathbf{x}_i-\mathbf{s}_j\|_2
    \label{eq:error_metric}
\end{equation}
where $\mathbf{S}=\{\mathbf{s}_j\in S_\theta|j = 1,2,...,M\}$ is a set of points densely and evenly sampled (with an interval of 0.02 in the synthetic experiments) on the superquadric surface $S_\theta$.
\begin{figure} [!htp]
    \centering
    \includegraphics[width=0.95\columnwidth]{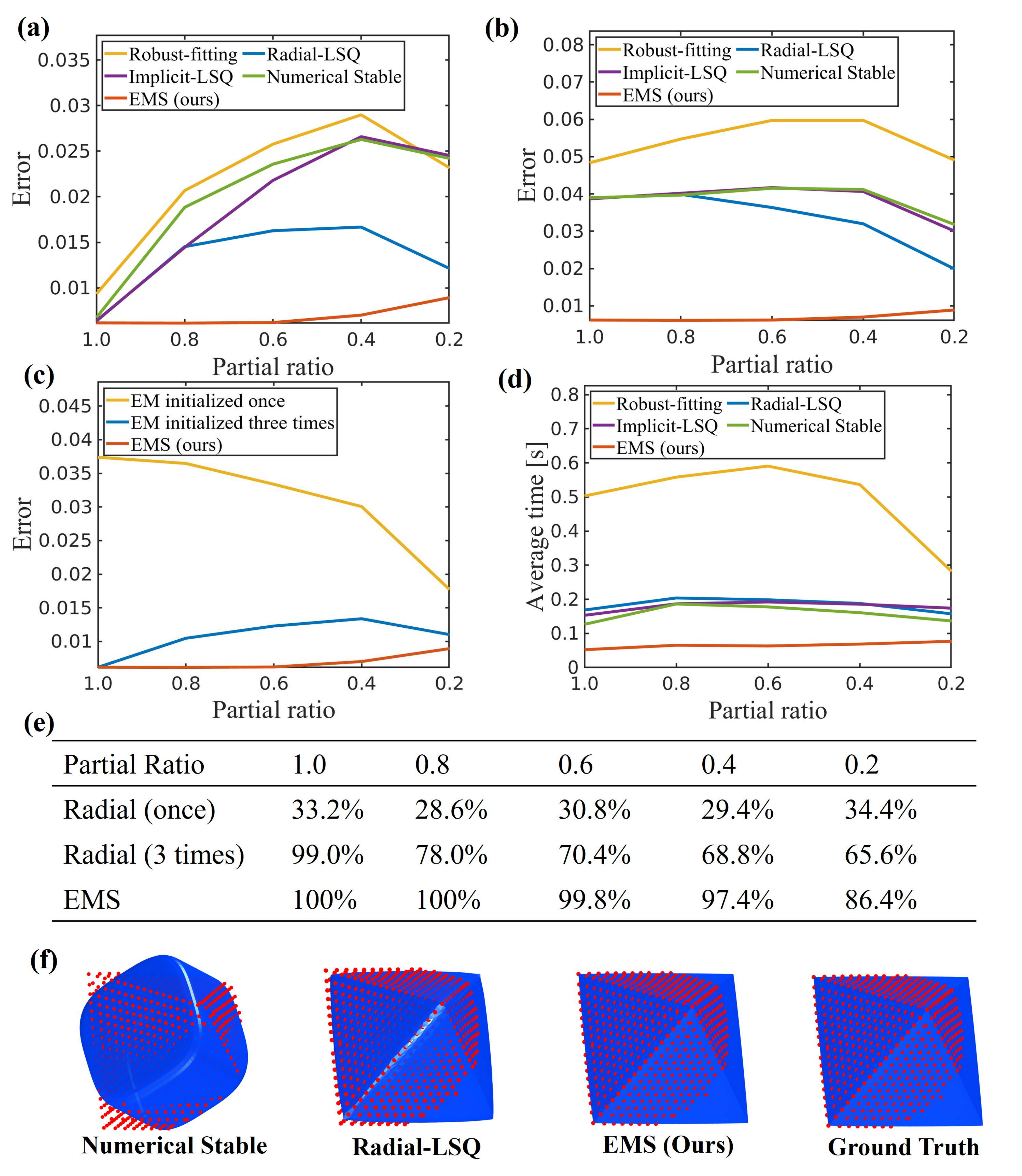} 
    \caption{Results of recovery from partial data. (a) Average error (the baselines are initialized 3 times). (b) Average error (the baselines are initialized once). (c) Ablation study. (d) Average runtime. (e) Global Optimality rate (declared when the fitting error is less than 0.01) (f) Examples at partial ratio 0.4, when the baselines get stuck in local optima, while ours have managed to jump out.}
    \label{fig:partial}
    \vspace{-0.3cm}
\end{figure}

\textbf{Recovery on partial data:} This experiment simulates the cases where points are sampled from a partial area of a superquadric in the sense that sometimes we can only obtain a partial view of an object.
For each point cloud of the 500 superquadrics, a point is randomly selected and a certain percentage (partial ratio) of the points which are the closest to the random point are kept.
We test on 5 different levels of partial ratio ranging from 1.0 (complete point cloud) to 0.2, resulting in totally 2500 point clouds. 
Results are shown in Fig. \ref{fig:partial}.
Our method significantly outperforms the baseline methods at all levels of partial ratios.
All the methods perform well when a complete point cloud of a superquadric is available.
However, with the decreasing of the partial ratio, it becomes harder to make a reasonable initial guess of the parameter.
Therefore, the baseline methods become more likely to get stuck at a local optimum, even with multiple trials.
While, our method continuously avoids the local optima with the geometry-guided switching strategy, and thus maintains a high success rate in converging to the global optimum.
We also perform ablation studies about the S-step.
The results verify its effectiveness (Fig. \ref{fig:partial}(c)).
Although our algorithm only guarantees local optimality, the empirical results indicate that our method is very likely to achieve the global optimum in practice (Fig. \ref{fig:partial}(e)).

\begin{figure} [!ht]
    \centering
    \includegraphics[width=0.90\columnwidth]{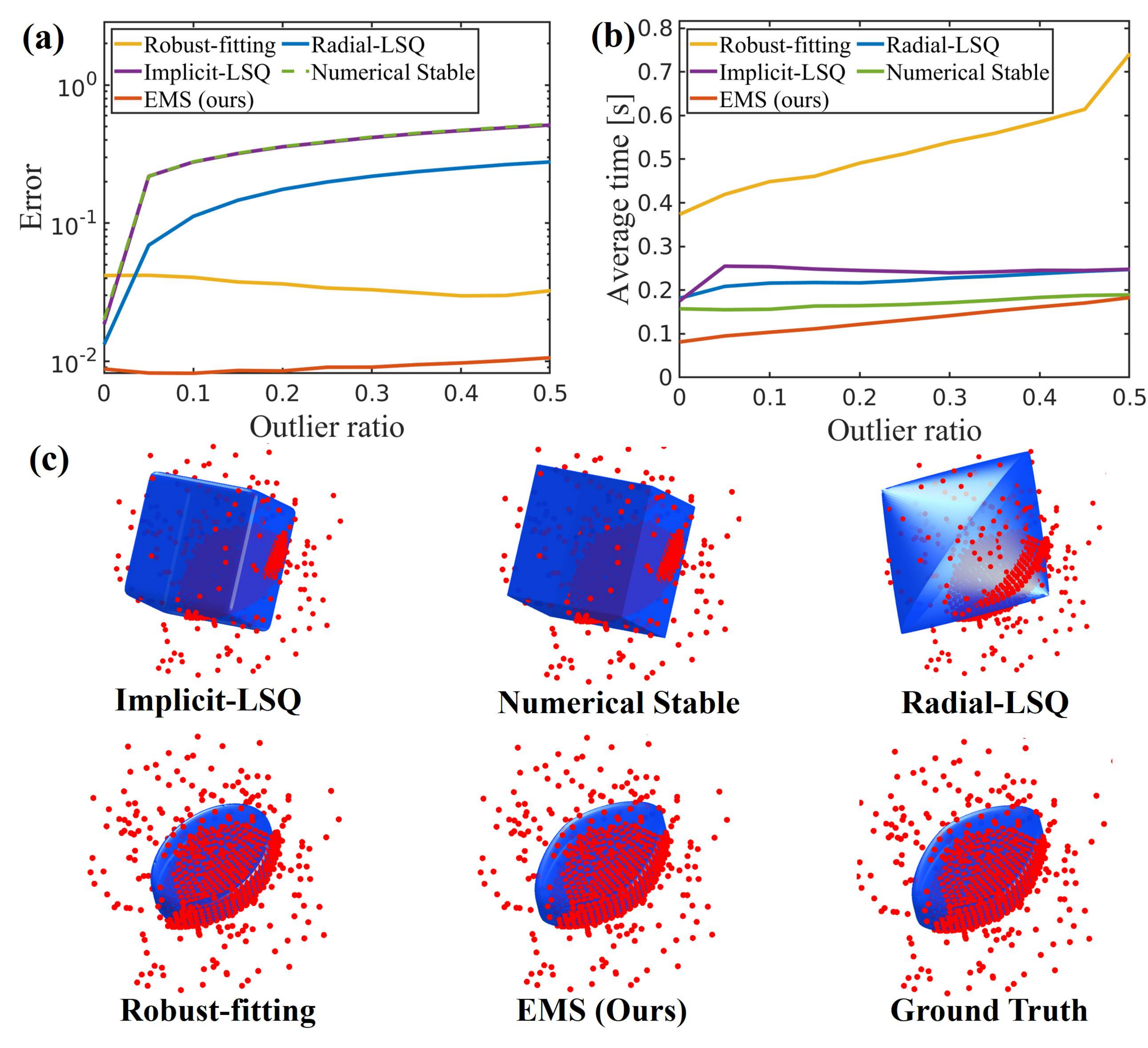} 
    \caption{Outlier results. (a) Average error at different outlier ratios. (b) Average runtime. Runtime of Radial, Implicit and NS has no significant meaning since they cannot recover the superquadric correctly in most cases. (c) Examples at outlier ratio 0.4.}
    \label{fig:outlier}
    \vspace{-0.2cm}
\end{figure}

\textbf{Robustness to outliers:} We add different levels of Gaussian outliers to the 2500 point clouds.
Outlier ratio (the number of outliers / the number of inliers) is used to quantify the level of outliers added.
Results are shown in Fig. \ref{fig:outlier}.
Our method shows superior robustness to outliers and is able to recover the underlying superquadrics even from severely corrupted point clouds.
The Implicit-LSQ, Radial-LSQ and Numerical Stable are vulnerable to outliers, since they are not designed to tolerate outliers.
The Robust-fitting performs better, however, is limited by its heuristic outlier rejection policy, which also imposes an undesirable negative effect on clean point clouds.
In contrast, our method infers outliers in a posterior probabilistic approach and thus maintains high robustness among all levels of outliers.

\begin{figure} [!ht]
    \centering
    \includegraphics[width=0.90\columnwidth]{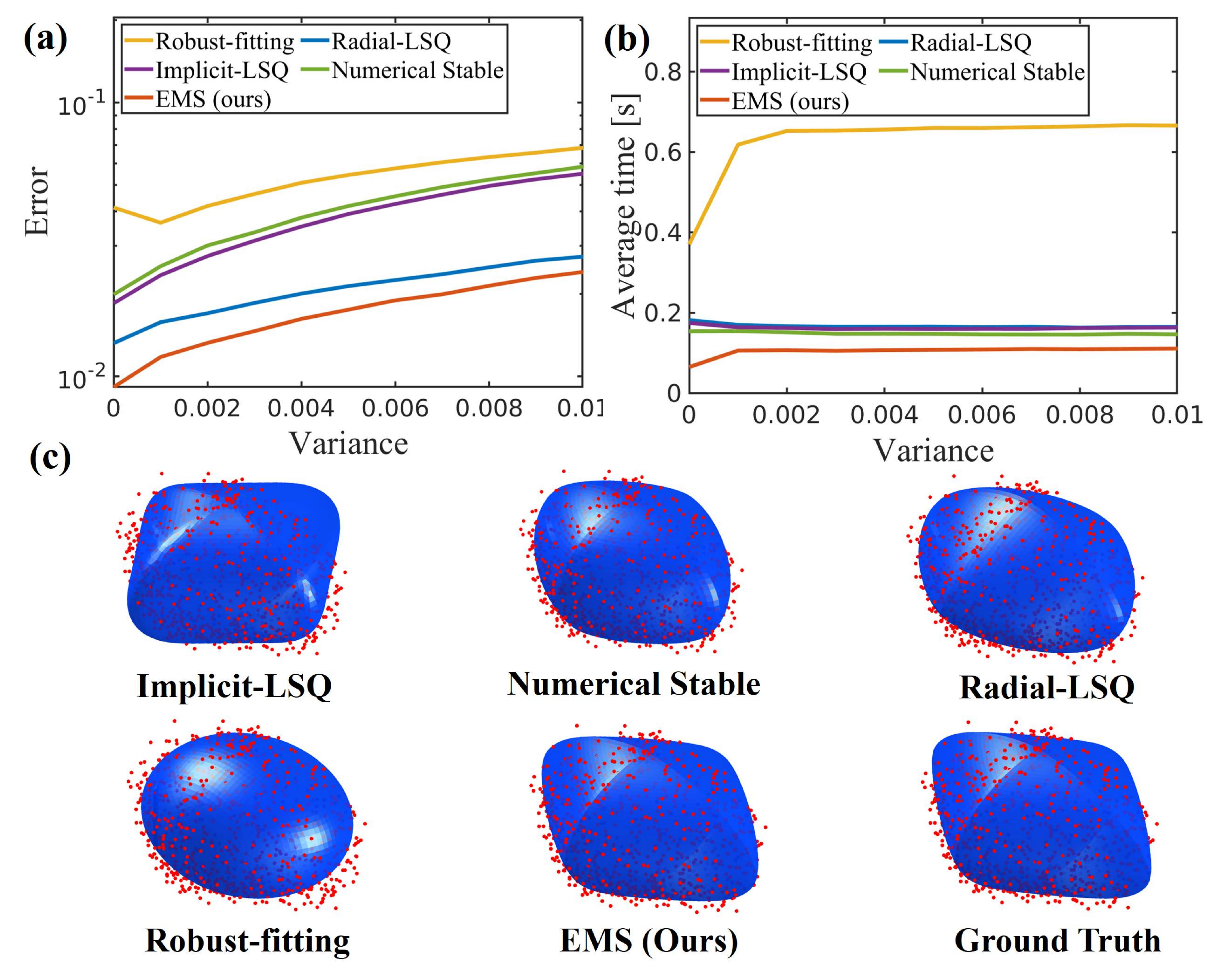} 
    \caption{Noise results (a) Average errors at different noise levels (evaluated on uncorrupted point clouds). (b) Average runtime. (c) Examples at $\sigma^2=0.01$ Gaussian noise level.}
    \label{fig:noise}
    \vspace{-0.6cm}
\end{figure}

\textbf{Robustness to noise:} 
In this experiment, we corrupt each point in the point clouds with Gaussian noise of different variance.
Results are shown in Fig. \ref{fig:noise}.
It can be observed that the fitting error grows with the increase of noise level for all the methods, including ours.
This is because, unlike outliers, the noise inherently corrupts the geometric information.
But still, our method shows advantages in robustness and efficiency when dealing with noisy point clouds.
\subsection{Recovery on Real-world Datasets}
\label{experiment_real}
In this section, we evaluate the single superquadric recovery on two public datasets: the KIT ObjectModel\cite{kasper2012kit} and BigBIRD\cite{singh2014bigbird}.
Both of the datasets consist of various daily household objects, and are widely used as benchmarks in 3D-object reconstruction, recognition, and robot grasping.
KIT is scanned by a high-accuracy laser scanner (Minolta VI-900), while BigBIRD is captured by a commercial-grade RGB-D camera (PrimeSense Carmine 1.08).

\begin{figure} [!htp]
    \centering
    \includegraphics[width=0.9\columnwidth]{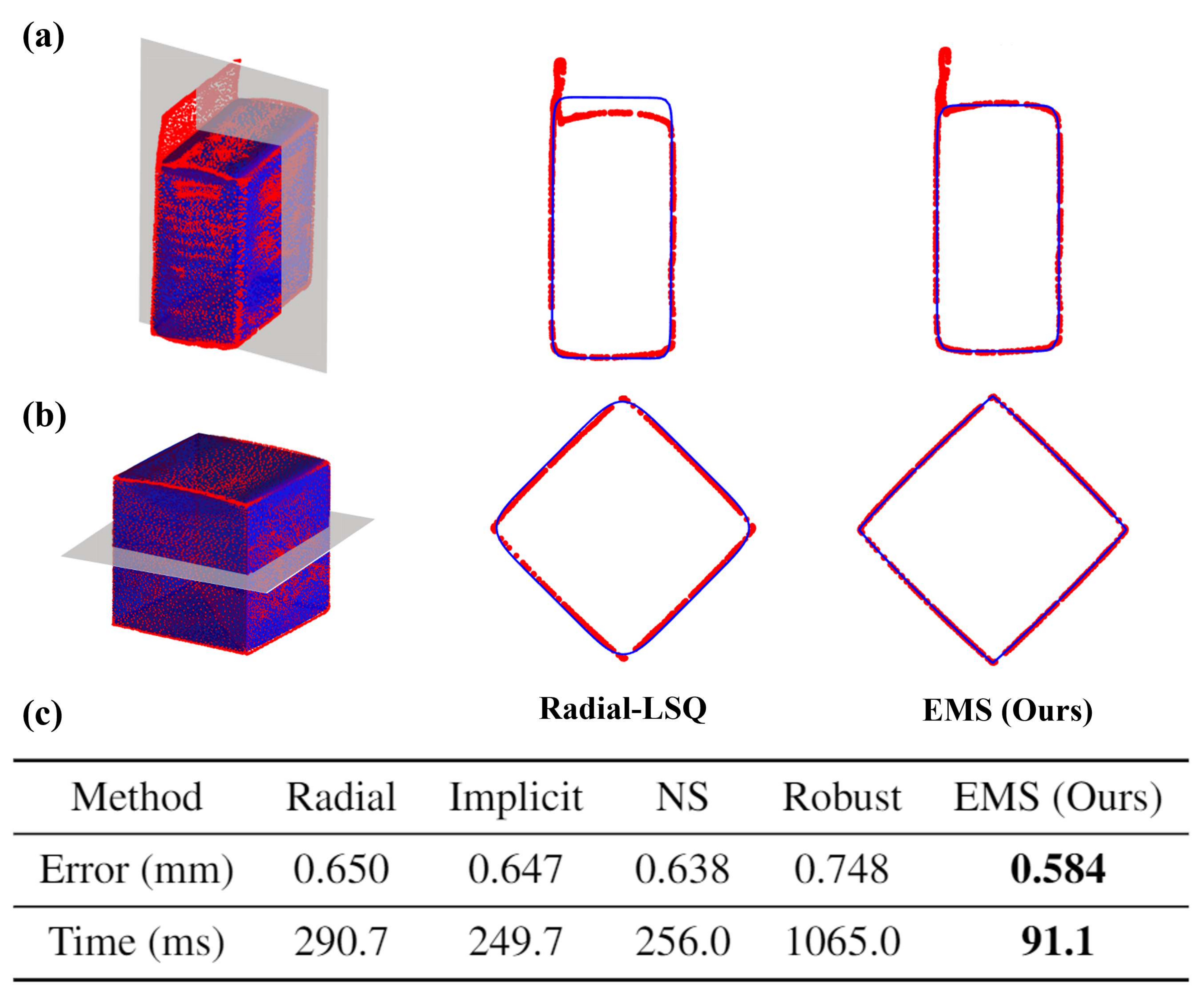} 
    \caption{KIT results. (a) and (b) are the two cross-section views of fitting results between Radial-LSQ and the proposed method. (c) Quantitative results of average error (Eq. \eqref{eq:error_metric} with sampling interval as dense as 0.1 mm) and runtime.}
    \label{fig:kit}
    \vspace{-0.3cm}
\end{figure}

\textbf{KIT ObjectModels:} 
We follow the setting in\cite{superquadrics_num_stable}.
The dataset includes 145 objects, among which \cite{superquadrics_num_stable} selects 105 items that can be reasonably approximated by a single superquadric. 
However, we decide to shrink our scope to 97 items, since some of the objects (\eg, bottles) would be better recovered with multiple superquadrics.
Therefore, we defer the discussion of those cases to Sec. \ref{sec:multi-superquadrics}.
We downsample the point clouds with voxelized grids to about 1500 points.
Quantitative results are shown in Fig. \ref{fig:kit}(c).
Our method outperforms all the baseline methods in accuracy and efficiency.
Our method also demonstrates some interesting features which can not be shown quantitatively.
As shown in Fig. \ref{fig:kit}(a), some objects have minor local deformations and/or parts deviating from the main structure, which affect the major shape capturing.
Our probabilistic method is able to reduce the impact of those parts (inferred as outliers) and achieves an overall better fitting.
Also in Fig. \ref{fig:kit}(b), the box-like objects are usually filled with content inside, resulting in a shape slightly extended in the middle.
This shape is within the vocabulary of superquadrics, however, is so close to cuboids that the other algorithms fail in telling them apart, \textit{i.e.}, get stuck at a local optimum. 
In contrast, our method is able to switch and check among similar shapes and recover the optimal superquadric.
\begin{figure} [!ht]
    \centering
    \includegraphics[width=0.9\columnwidth]{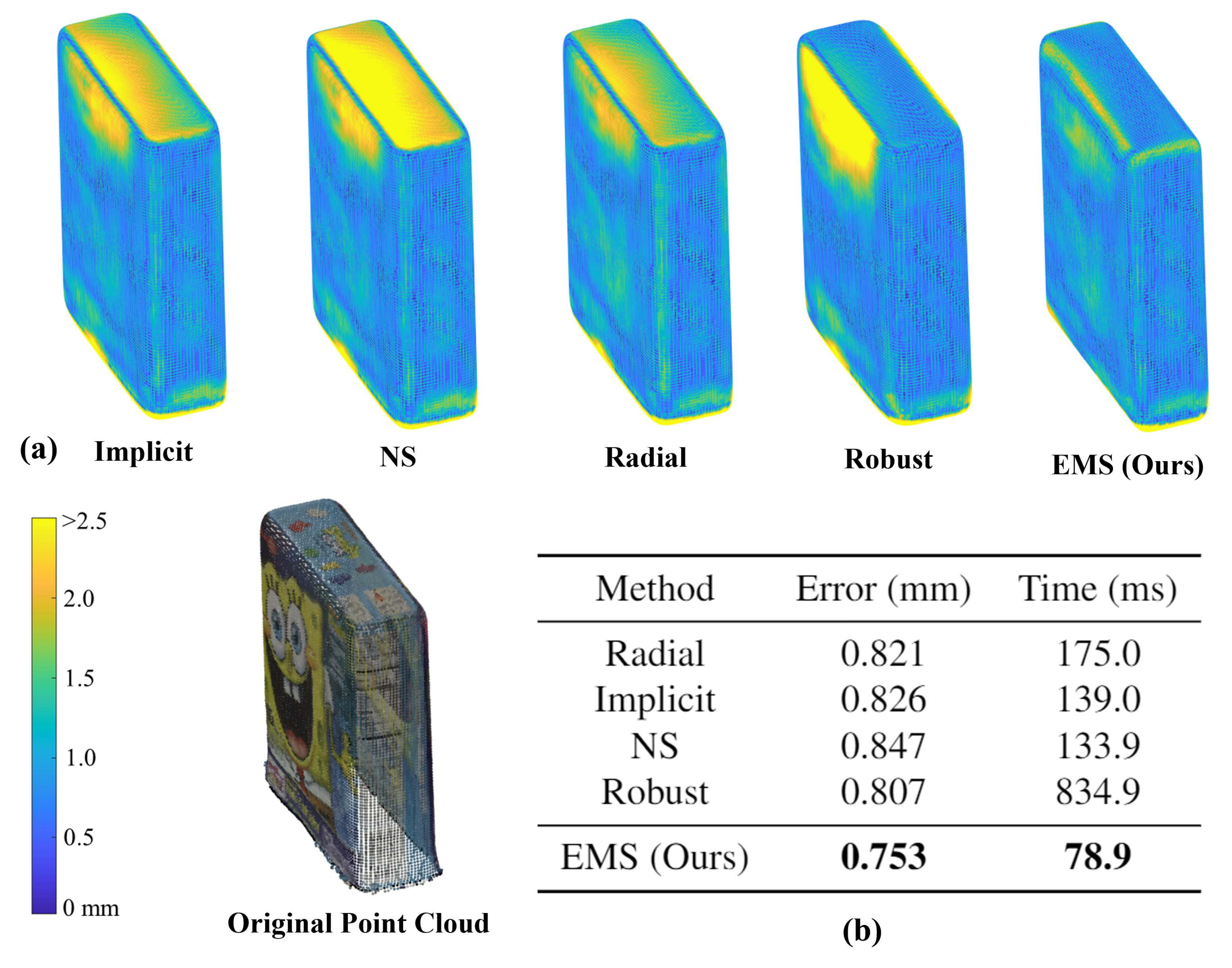} 
    \caption{BigBIRD results. (a) A heatmap showing the local fitting error between the recovered superquadric and the original point cloud. (b) Quantitative results of the average error (Eq. \eqref{eq:error_metric} with sampling interval as dense as 0.1 mm) and runtime.}
    \label{fig:bigbird}
    \vspace{-0.4cm}
\end{figure}

\textbf{BigBIRD:}
Compared with laser scanners, low-cost RGB-D cameras are inferior in measurement accuracy, resulting in noisy point clouds with outliers.
Therefore, the BigBIRD is an ideal dataset for evaluating the robustness of an algorithm in daily operation.
The dataset consists of 125 object instances, and we test on 91 of them which can be reasonably approximated with single superquadrics.
The point clouds are downsampled with voxelized grids to around 1000 points.
Results are shown in Fig. \ref{fig:bigbird}.
Among the baselines, the Robust-fitting shows better robustness but takes a much longer time to converge.
Our method can achieve the best accuracy with the least time consumption.
\begin{figure*}[!htp]
    \centering 
    \includegraphics[scale=0.064]{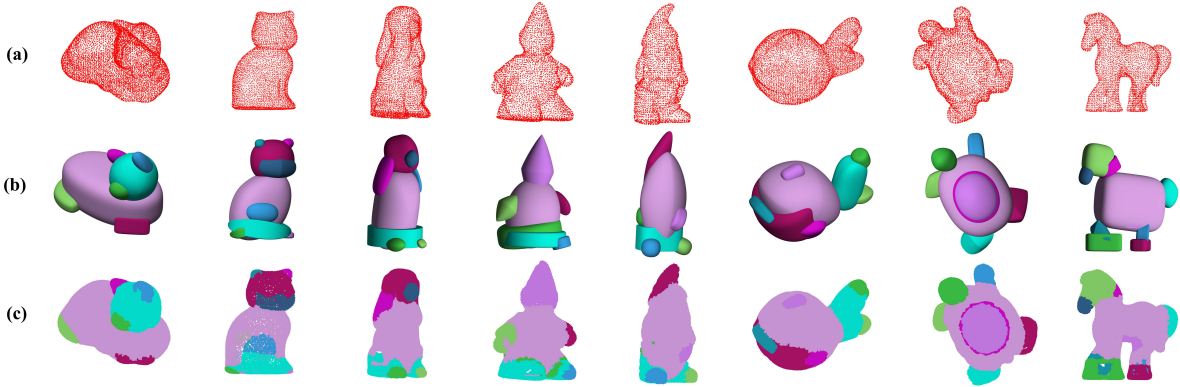} 
    \caption{Examples of multi-superquadrics recovery. (a) The original point clouds. (b) The recovered multi-superquadrics representations with the proposed method. (c) Segmentation of the original point clouds based on the multi-superquadrics representations.}
    \label{fig:multi_complex}
\vspace{-0.4cm}
\end{figure*}

\subsection{Multi-superquadrics Recovery}
\label{sec:multi-superquadrics}
\begin{figure}[!ht]
    \centering 
    \includegraphics[width=0.80\columnwidth]{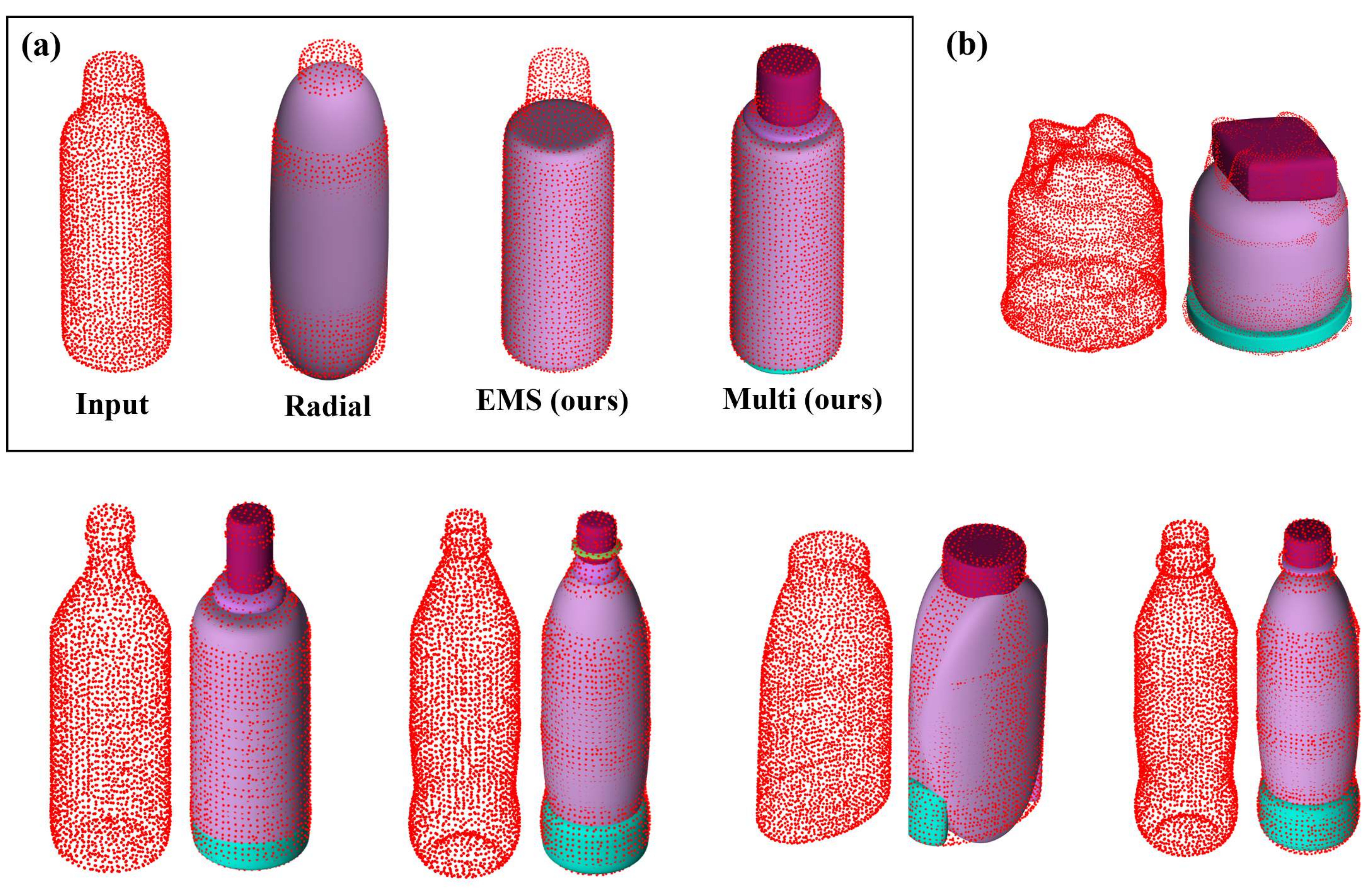} 
    \caption{Qualitative multi-superquadrics recovery on bottles from KIT. (a) Comparison of single superquadric recovery between Radial-LSQ and EMS (ours). The proposed method is able to recover the major superquadric shape (a cylinder) from the point cloud. (b) Other results of the proposed method.}
    \label{fig:bottles}
\vspace{-0.5cm}
\end{figure}

In this section, we show the qualitative results of our extension to the multi-superquadrics recovery (Sec. \ref{sec:extension}).
We use the complex objects which cannot be properly approximated by a single superquadric from the KIT ObjectModels.
We downsample the point clouds to around 5000 points, so as to keep enough details of the complex shapes.
We set the maximum hierarchical layers to 3 and the pruning threshold to 60 points.
As shown in Fig. \ref{fig:bottles}, the baseline single superquadric method fails in capturing the major superquadric-like part of a bottle as a cylinder.
Our method succeeds in doing this and identifies the points around the cap of the bottle as outliers, from which the bottle is recovered hierarchically. 
Other results on more complex objects are shown in Fig. \ref{fig:multi_complex}.
Our method is able to represent the objects vividly with superquadrics.
Some detailed parts such as ears and mouths of animals can also be recovered.
\section{Discussion \& Limitations}
Various experiments show that our method is robust, accurate and efficient.
As shown in Sec. \ref{experiment_synthetic} and Sec. \ref{experiment_real}, the baseline methods are sensitive to initialization, and are easily trapped into local optima. 
Therefore, to obtain a satisfying result, they usually require multiple trials starting from different initial guesses.
In comparison, our method is able to avoid local optima on the fly and achieve a better recovery accuracy with one shot.
Also, our method is robust to outliers and noise.
This feature is brought in by our probabilistic formulation of the problem, where we explicitly model the possible sources of outliers and noise with a uniform distribution and a Gaussian distribution, respectively.
This feature also forms the basis for our multi-superquadrics extension.
As shown in Sec. \ref{sec:multi-superquadrics}, our method is able to capture the major superquadric-like part from a point cloud and hierarchically generate subordinate superquadrics from the `outliers' identified at the upper level.

However, there is a limitation in our multi-superquadrics extension: our approach requires the object being recovered possesses an inherent hierarchical geometric structure.
For more general cases, a preliminary point cloud segmentation is preferred \cite{qi2017pointnet, qi2017pointnetplusplus}.
Another limitation is due to the expressiveness of superquadrics.
It is hard to well capture the shape of a thin-walled non-convex object such as a cup.
In future works, we plan to adapt deformations and primitive subtractions to extend the vocabulary of superquadrics.

\section{Conclusion}
We present the first probabilistic method to recover superquadrics from point clouds. 
We formulate the recovering process as an MLE problem and propose a novel geometry-guided approach, EMS, to infer an optimal superquadric. 
The proposed method outperforms the state-of-the-art in terms of accuracy, robustness, and efficiency, which is substantiated by extensive experiments on both synthetic and real-world datasets. 
We also extend our method to represent a complex object with a hierarchy of superquadrics.
Such representation can reveal the geometric and hierarchical structure of an object, which, we believe, has great potential in facilitating tasks such as object classification, segmentation and grasping.

\vspace{\baselineskip}
\noindent \textbf{Acknowledgements}
This research is supported by the National Research Foundation, Singapore, under its Medium Sized Centre Programme - Centre for Advanced Robotics Technology Innovation (CARTIN) R-261-521-002-592, and JHU internal funds.



{\small
\bibliographystyle{ieee_fullname}
\bibliography{main}

\begin{thebibliography}{10}\itemsep=-1pt

\bibitem{Bajcsy1987ICCV}
R. Bajcsy and F. Solina.
\newblock Three dimensional object representation revisited.
\newblock In {\em Proceedings of the IEEE/CVF International Conference on
  Computer Vision (ICCV)}, pages 231--240, 1987.

\bibitem{barr1981superquadrics}
A.~H. Barr.
\newblock Superquadrics and angle-preserving transformations.
\newblock {\em IEEE Computer graphics and Applications}, 1(1):11--23, 1981.

\bibitem{biederman1987recognition}
I. Biederman.
\newblock Recognition-by-components: a theory of human image understanding.
\newblock {\em Psychological review}, 94(2):115, 1987.

\bibitem{ObjectDetectionICRA}
G. Biegelbauer and M. Vincze.
\newblock Efficient {3D} object detection by fitting superquadrics to range
  image data for robot's object manipulation.
\newblock In {\em Proceedings 2007 IEEE International Conference on Robotics
  and Automation (ICRA)}, pages 1086--1091, 2007.

\bibitem{BORGES1997655}
D.~L. Borges and R.~B. Fisher.
\newblock Class-based recognition of 3d objects represented by volumetric
  primitives.
\newblock {\em Image and Vision Computing}, 15(8):655--664, 1997.
\newblock British Machine Vision Conference.

\bibitem{chevalier2003segmentation}
L. Chevalier, J. Jaillet, and A. Baskurt.
\newblock Segmentation and superquadric modeling of {3D} objects.
\newblock In {\em The 11-th International Conference in Central Europe on
  Computer Graphics, Visualization and Computer Vision'2003, {WSCG} 2003},
  2003.

\bibitem{reflective94}
T.~F. Coleman and Y. Li.
\newblock On the convergence of reflective newton methods for large-scale
  nonlinear minimization subject to bounds.
\newblock {\em Math. Program.}, 67:189--224, 10 1994.

\bibitem{trust_region_reflective}
T.~F. Coleman and Y. Li.
\newblock An interior trust region approach for nonlinear minimization subject
  to bounds.
\newblock {\em SIAM Journal on Optimization}, 6(2):418--445, 1996.

\bibitem{voxel96}
B. Curless and M. Levoy.
\newblock A volumetric method for building complex models from range images.
\newblock In {\em Proceedings of the 23rd Annual Conference on Computer
  Graphics and Interactive Techniques}, SIGGRAPH '96, page 303–312, New York,
  NY, USA, 1996. Association for Computing Machinery.

\bibitem{DEMP1977}
A.~P. Dempster, N.~M. Laird, and D.~B. Rubin.
\newblock Maximum likelihood from incomplete data via the {EM} algorithm.
\newblock {\em Journal of the Royal Statistical Society: Series B
  (Methodological)}, 39(1):1--22, 1977.

\bibitem{fan2017a}
H. Fan, H. Su, and L.~J. Guibas.
\newblock A point set generation network for {3D} object reconstruction from a
  single image.
\newblock In {\em Proceedings of the IEEE Conference on Computer Vision and
  Pattern Recognition (CVPR)}, July 2017.

\bibitem{GrossICCV88}
A.~D. Gross and T.~E. Boult.
\newblock Error of fit measures for recovering parametric solids.
\newblock In {\em Proceedings of the IEEE/CVF International Conference on
  Computer Vision (ICCV)}, pages 690--694, 1988.

\bibitem{hu1995robust}
Y. Hu and W.~G. Wee.
\newblock {Robust 3D part extraction from range images with deformable
  superquadric models}.
\newblock In {\em Signal Processing, Sensor Fusion, and Target Recognition IV},
  volume 2484, pages 524 -- 535. International Society for Optics and
  Photonics, SPIE, 1995.

\bibitem{kasper2012kit}
A. Kasper, Z. Xue, and R. Dillmann.
\newblock The {KIT} object models database: An object model database for object
  recognition, localization and manipulation in service robotics.
\newblock {\em The International Journal of Robotics Research}, 31(8):927--934,
  2012.

\bibitem{core_systems_in_human_cognition}
D.~K. Katherine and S.~S. Elizabeth.
\newblock Core systems in human cognition.
\newblock In {\em From Action to Cognition}, volume 164 of {\em Progress in
  Brain Research}, pages 257--264. Elsevier, 2007.

\bibitem{leonardis1997superquadrics}
A. Leonardis, A. Jaklic, and F. Solina.
\newblock Superquadrics for segmenting and modeling range data.
\newblock {\em IEEE Transactions on Pattern Analysis and Machine Intelligence},
  19(11):1289--1295, 1997.

\bibitem{levenberg1944method}
K. Levenberg.
\newblock A method for the solution of certain non-linear problems in least
  squares.
\newblock {\em Quarterly of Applied Mathematics}, 2(2):164--168, 1944.

\bibitem{lorensen1987marching}
W.~E. Lorensen and H.~E. Cline.
\newblock Marching cubes: A high resolution {3D} surface construction
  algorithm.
\newblock {\em SIGGRAPH Comput. Graph.}, 21(4):163–169, Aug. 1987.

\bibitem{MatlabOTB}
Matlab optimization toolbox, 2021a.
\newblock The MathWorks, Natick, MA, USA.

\bibitem{cube_cvpr2018}
C. Niu, J. Li, and K. Xu.
\newblock {Im2Struct}: Recovering {3D} shape structure from a single {RGB}
  image.
\newblock In {\em Proceedings of the IEEE Conference on Computer Vision and
  Pattern Recognition (CVPR)}, June 2018.

\bibitem{Oblak_access}
T. Oblak, J. Šircelj, V. Štruc, P. Peer, F. Solina, and A. Jaklič.
\newblock Learning to predict superquadric parameters from depth images with
  explicit and implicit supervision.
\newblock {\em IEEE Access}, 9:1087--1102, 2021.

\bibitem{paschalidou2020learning}
D. Paschalidou, L.~V. Gool, and A. Geiger.
\newblock Learning unsupervised hierarchical part decomposition of {3D} objects
  from a single {RGB} image.
\newblock In {\em Proceedings of the IEEE/CVF Conference on Computer Vision and
  Pattern Recognition (CVPR)}, June 2020.

\bibitem{paschalidou2019superquadrics}
D. Paschalidou, A.~O. Ulusoy, and A. Geiger.
\newblock Superquadrics revisited: Learning {3D} shape parsing beyond cuboids.
\newblock In {\em Proceedings of the IEEE/CVF Conference on Computer Vision and
  Pattern Recognition (CVPR)}, June 2019.

\bibitem{pentland1987perceptual}
A.~P. Pentland.
\newblock Perceptual organization and the representation of natural form.
\newblock In M.~A. Fischler and O. Firschein, editors, {\em Readings in
  Computer Vision}, pages 680--699. Morgan Kaufmann, San Francisco (CA), 1987.

\bibitem{qi2017pointnet}
C.~R. Qi, H. Su, K. Mo, and L.~J. Guibas.
\newblock {PointNet}: Deep learning on point sets for {3D} classification and
  segmentation.
\newblock In {\em Proceedings of the IEEE Conference on Computer Vision and
  Pattern Recognition (CVPR)}, July 2017.

\bibitem{qi2017pointnetplusplus}
C.~R. Qi, L. Yi, H. Su, and L.~J. Guibas.
\newblock {PointNet++}: Deep hierarchical feature learning on point sets in a
  metric space.
\newblock In {\em Proceedings of the 31st International Conference on Neural
  Information Processing Systems}, NIPS'17, page 5105–5114, Red Hook, NY,
  USA, 2017. Curran Associates Inc.

\bibitem{quispe2015exploiting}
A.~H. Quispe, B. Milville, M.~A. Gutiérrez, C. Erdogan, M. Stilman, H.
  Christensen, and H.~B. Amor.
\newblock Exploiting symmetries and extrusions for grasping household objects.
\newblock In {\em 2015 IEEE International Conference on Robotics and Automation
  (ICRA)}, pages 3702--3708, 2015.

\bibitem{RUAN_closedform}
S. Ruan and G.~S. Chirikjian.
\newblock Closed-form {Minkowski} sums of convex bodies with smooth positively
  curved boundaries.
\newblock {\em Computer-Aided Design}, 143:103133, 2022.

\bibitem{ruan2019efficient}
S. Ruan, K.~L. Poblete, Y. Li, Q. Lin, Q. Ma, and G.~S. Chirikjian.
\newblock Efficient exact collision detection between ellipsoids and
  superquadrics via closed-form {Minkowski} sums.
\newblock In {\em 2019 International Conference on Robotics and Automation
  (ICRA)}, pages 1765--1771, 2019.

\bibitem{singh2014bigbird}
A. Singh, J. Sha, K.~S. Narayan, T. Achim, and P. Abbeel.
\newblock {BigBIRD}: A large-scale {3D} database of object instances.
\newblock In {\em 2014 IEEE International Conference on Robotics and Automation
  (ICRA)}, pages 509--516, 2014.

\bibitem{sirceljsegmentation}
J. {\v{S}}ircelj, T. Oblak, K. Grm, U. Petkovi{\'c}, A. Jakli{\v{c}}, P. Peer,
  V. {\v{S}}truc, and F. Solina.
\newblock Segmentation and recovery of superquadric models using convolutional
  neural networks.
\newblock In {\em 25th Computer Vision Winter Workshop}, 2020.

\bibitem{solina1990recovery}
F. Solina and R. Bajcsy.
\newblock Recovery of parametric models from range images: the case for
  superquadrics with global deformations.
\newblock {\em IEEE Transactions on Pattern Analysis and Machine Intelligence},
  12(2):131--147, 1990.

\bibitem{Tulsiani_2017_CVPR}
S. Tulsiani, H. Su, L.~J. Guibas, A.~A. Efros, and J. Malik.
\newblock Learning shape abstractions by assembling volumetric primitives.
\newblock In {\em Proceedings of the IEEE Conference on Computer Vision and
  Pattern Recognition (CVPR)}, July 2017.

\bibitem{vanDop98CVPR}
E.~R. Van~Dop and P.~P.~L. Regtien.
\newblock Fitting undeformed superquadrics to range data: improving model
  recovery and classification.
\newblock In {\em Proceedings of IEEE Computer Society Conference on Computer
  Vision and Pattern Recognition (CVPR)}, pages 396--401, 1998.

\bibitem{superquadrics_num_stable}
N. Vaskevicius and A. Birk.
\newblock Revisiting superquadric fitting: A numerically stable formulation.
\newblock {\em IEEE Transactions on Pattern Analysis and Machine Intelligence},
  41(1):220--233, 2019.

\bibitem{vezzani2017grasping}
G. Vezzani, U. Pattacini, and L. Natale.
\newblock A grasping approach based on superquadric models.
\newblock In {\em 2017 IEEE International Conference on Robotics and Automation
  (ICRA)}, pages 1579--1586, 2017.

\bibitem{vezzani2018improving}
G. Vezzani, U. Pattacini, G. Pasquale, and L. Natale.
\newblock Improving superquadric modeling and grasping with prior on object
  shapes.
\newblock In {\em 2018 IEEE International Conference on Robotics and Automation
  (ICRA)}, pages 6875--6882, 2018.

\bibitem{zeng20163dmatch}
A. Zeng, S. Song, M. Niessner, M. Fisher, J. Xiao, and T. Funkhouser.
\newblock {3DMatch}: Learning local geometric descriptors from {RGB-D}
  reconstructions.
\newblock In {\em Proceedings of the IEEE Conference on Computer Vision and
  Pattern Recognition (CVPR)}, July 2017.

\bibitem{cube_cvpr2017}
C. Zou, E. Yumer, J. Yang, D. Ceylan, and D. Hoiem.
\newblock {3D-PRNN}: Generating shape primitives with recurrent neural
  networks.
\newblock In {\em Proceedings of the IEEE International Conference on Computer
  Vision (ICCV)}, Oct 2017.

\end{thebibliography}


\begin{thebibliography}{1}\itemsep=-1pt

\bibitem{trust_region_reflective}
T.~F. Coleman and Y. Li.
\newblock An interior trust region approach for nonlinear minimization subject
  to bounds.
\newblock {\em SIAM Journal on Optimization}, 6(2):418--445, 1996.

\bibitem{GrossICCV88}
A.~D. Gross and T.~E. Boult.
\newblock Error of fit measures for recovering parametric solids.
\newblock In {\em Proceedings of the IEEE/CVF International Conference on
  Computer Vision (ICCV)}, pages 690--694, 1988.

\bibitem{hu1995robust}
Y. Hu and W.~G. Wee.
\newblock {Robust 3D part extraction from range images with deformable
  superquadric models}.
\newblock In {\em Signal Processing, Sensor Fusion, and Target Recognition IV},
  volume 2484, pages 524 -- 535. International Society for Optics and
  Photonics, SPIE, 1995.

\bibitem{MatlabOTB}
Matlab optimization toolbox, 2021a.
\newblock The MathWorks, Natick, MA, USA.

\bibitem{pilu1995equal}
M. Pilu and R.~B. Fisher.
\newblock Equal-distance sampling of superellipse models.
\newblock In {\em Proceedings of the 1995 British Conference on Machine Vision
  (Vol. 1)}, BMVC '95, page 257–266, GBR, 1995. BMVA Press.

\bibitem{solina1990recovery}
F. Solina and R. Bajcsy.
\newblock Recovery of parametric models from range images: the case for
  superquadrics with global deformations.
\newblock {\em IEEE Transactions on Pattern Analysis and Machine Intelligence},
  12(2):131--147, 1990.

\bibitem{superquadrics_num_stable}
N. Vaskevicius and A. Birk.
\newblock Revisiting superquadric fitting: A numerically stable formulation.
\newblock {\em IEEE Transactions on Pattern Analysis and Machine Intelligence},
  41(1):220--233, 2019.

\end{thebibliography}
}

\end{document}


\title{Supplementary Material of Robust and Accurate Superquadric Recovery: \\a Probabilistic Approach}

\author{Weixiao Liu$^{1, 2}$ \quad Yuwei Wu$^{1}$ \quad Sipu Ruan$^{1}$ \quad Gregory S. Chirikjian$^{1}$\footnotemark[1]\\
$^1$National University of Singapore \quad $^2$Johns Hopkins University\\
{\tt\small \{mpewxl, yw.wu, ruansp, mpegre\}@nus.edu.sg}
}
\maketitle

\begin{abstract}
   In this supplementary material, we provide the detailed derivations, discussions and experiment settings. In Sec. \ref{superquadric_area}, the approximation of the superquadric surface area is detailed. In Sec. \ref{optimization}, we further discuss the stability problem about the optimization. In Sec. \ref{similarity}, we present the mathematical formulation of the S-step, which includes the candidate generation and the switching strategy. In Sec. \ref{sampling}, we introduce how to conduct equal-distance sampling on a superquadric surface. Finally, in Sec. \ref{experiment}, detailed experiment settings are presented.
\end{abstract}

\section{Interpolation of Superquadric Area}
\label{superquadric_area}
Generally, the surface area $A_\theta$ of a superquadric cannot be expressed in closed-form with $\{\epsilon_1, \epsilon_2, a_x, a_y, a_z\}$ (note that the surface area is independent of the pose $\{\mathbf{R}, \mathbf{t}\}$).
However, closed-form solution is available when the shape parameter $\{\epsilon_1,\epsilon_2\}$ is a combination of 0 and 2.

(1) When $\{\epsilon_1, \epsilon_2\} = \{0, 0\}$, the superquadric is a cuboid, whose surface area is
\begin{equation}
    A_{\{0,0\}} = 8(a_x a_y + a_ya_z +a_x a_z)
\end{equation}

(2) When $\{\epsilon_1, \epsilon_2\} = \{0, 2\}$, the superquadric is a hexahedron, whose surface area is
\begin{equation}
    A_{\{0,2\}} = 8a_z(a_x^2 + a_y^2)^{\frac{1}{2}}+4a_x a_y
\end{equation}

(3) When $\{\epsilon_1, \epsilon_2\} = \{0, 2\}$, the superquadric is a octahedron, whose surface area is
\begin{equation}
    A_{\{2,0\}} = 4(a_x(a_y^2+a_z^2)^{\frac{1}{2}}+a_y(a_x^2+a_z^2)^{\frac{1}{2}})
\end{equation}

(4) Lastly, when $\{\epsilon_1, \epsilon_2\} = \{2, 2\}$, the superquadric is a octahedron, whose surface area is
\begin{equation}
    A_{\{2,2\}} = 8\left(a_{0}(a_{0}-a_{xy})(a_{0}-a_{yz})(a_{0}-a_{xz})\right)^{\frac{1}{2}}
\end{equation}
where
\begin{equation*}
\begin{aligned}
    &a_{xy} = (a_x^2+a_y^2)^{\frac{1}{2}}\\
    &a_{yz} = (a_y^2+a_z^2)^{\frac{1}{2}}\\
    &a_{xz} = (a_x^2+a_z^2)^{\frac{1}{2}}\\
    &a_{0} = (a_{xy}+a_{yz}+a_{xz})/2
\end{aligned}
\end{equation*}
We approximate the area with a bi-linear interpolation
\begin{equation}
    A_{\theta} = 
    \begin{bmatrix}
    1-\epsilon_1/2\\
    \epsilon_1/2
    \end{bmatrix}^T
    \begin{bmatrix}
    A_{\{0,0\}} & A_{\{0,2\}}\\
    A_{\{2,0\}} & A_{\{2,2\}}
    \end{bmatrix}
    \begin{bmatrix}
    1-\epsilon_2/2\\
    \epsilon_2/2
    \end{bmatrix}
\end{equation}
We test the interpolation method exhaustively throughout the convex region of superquadrics, and it shows an average relative error of less than $10\%$ compared with the the area calculated through the triangular mesh.
More importantly, the bi-linear interpolation can fully capture the property that the surface area of a superquadric grows monotonically with $\epsilon_1$ and $\epsilon_2$.

\section{Discussion on Optimization}
\label{optimization}
\begin{figure}[!h]
    \centering 
    \includegraphics[width=1\columnwidth]{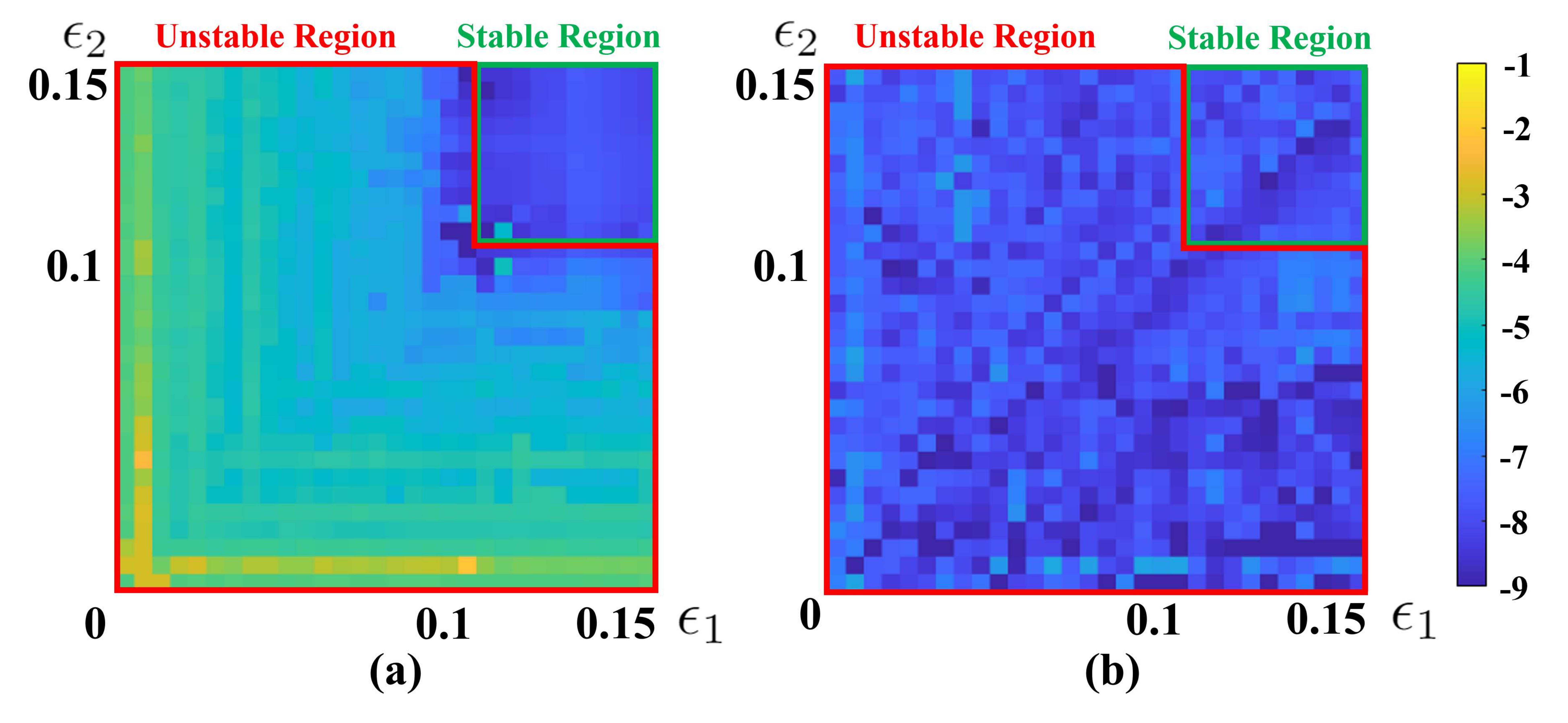}
    \caption{Performance in the unstable region. The error is evaluated by $\log_{10}(\|[\epsilon_1,\epsilon_2, a_x, a_y, a_z]-[\epsilon_1,\epsilon_2, a_x, a_y, a_z]_{gt}\|_2)$, where $[\epsilon_1,\epsilon_2, a_x, a_y, a_z]$ is the estimated shape and scale parameters and $[\epsilon_1,\epsilon_2, a_x, a_y, a_z]_{gt}$ is the ground-truth. (a) The recovery error of NS. (b) The recovery error of the proposed method.}
    \label{fig:stability}
\vspace{-0.2cm}
\end{figure}

The Levenberg–Marquardt (LM) algorithm is widely used in superquadrics fitting \cite{GrossICCV88,hu1995robust,solina1990recovery,superquadrics_num_stable}.
However, it has been confirmed that the optimization suffers from numerical instability as either one of the shape parameters $\epsilon_1$ or $\epsilon_2$ approaches 0.
As a consequence, most of the methods compromise by constraining the lower bounds of the shape parameters to 0.1, resulting in less accuracy when representing shapes with sharp edges, \textit{e.g.}, cuboids and cylinders.
In \cite{superquadrics_num_stable}, the authors claim that the problem is caused by the inherent instability of the implicit function and its gradient.
They solve this problem by approximating the implicit function with linear functions in the unstable region. 
We find out that the instability can also be explained as being introduced by the way that the LM deals with the bound.
When the optimization steps outside of the bound, it will be simply projected back to the nearest point on the bound.
Therefore, if we set the lower bounds of $\epsilon_1$ and $\epsilon_2$ to 0, the LM algorithm is likely to be forced to visit the points on the hyperplane of the lower bounds, where the implicit function is not well defined.
In contrast, the trust-region-reflective \cite{trust_region_reflective} tackles the bounding condition differently by conducting a line search along the reflective path, and thus avoid directly checking the value of the implicit function on the bound.
Therefore, by utilizing the trust-region-reflective, our method can maintain numerical stability without approximating the implicit function, as shown in Fig. \ref{fig:stability}.
It can be observed that both of the methods are stable; however, \cite{superquadrics_num_stable} shows less accuracy in recovering the superquadric parameter within the unstable region, which is caused by their approximation of the implicit function.

\section{Similarity and Switching}
\label{similarity}
In this section, we show the detailed mathematics formulation of the S-step (geometry-guided local optimum avoidance) in Sec. 3.5 of the paper. 
First, we formulate how to search for candidate parameters $\{\boldsymbol{\theta}^c_i\}$ encoding similar superquadrics from the current estimation $\boldsymbol{\theta}$, which is, presumably, a local optimum.
Recall that the superquadric parameter $\boldsymbol{\theta} = \{\epsilon_1, \epsilon_2, a_x, a_y, a_z, \mathbf{R}, \mathbf{t}\}$ (Sec. 2.1 in the papaer).
We denote $\mathbf{R} = [\mathbf{r}_1, \mathbf{r}_2, \mathbf{r}_3]$, where $\mathbf{r}_1$, $\mathbf{r}_2$ and $\mathbf{r}_3$ are orthonormal column vectors, corresponding to the directions of the x, y and z-axis (principal axis) of the superquadric frame, respectively.

\begin{figure}[!h]
    \centering 
    \includegraphics[width=0.9\columnwidth]{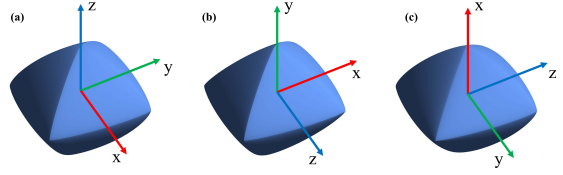}
    \caption{Similar superquadrics encoded by candidate parameters generated via axis-mismatch similarity. (a) The underlying shape of the current estimation $\boldsymbol{\theta}=[1.5, 1.7, 1, 1.2, 0.8, \mathbf{I}, [0, 0, 0]^T]$. (b) The underlying shape of the candidate $\boldsymbol{\theta}^c_1$. (c) The underlying shape of the candidate $\boldsymbol{\theta}^c_2$.}
    \label{fig:axis_mismatch}
\vspace{-0.2cm}
\end{figure}

\textbf{Axis-mismatch similarity}: 
In this category, we can obtain 2 candidates by re-assigning the principal axis to the x-axis and the y-axis, and re-arranging the corresponding shape, scale and rotation parameters:
\begin{equation}
\label{eq:Axis-mismatch}
\begin{aligned}
    \boldsymbol{\theta}^c_1 = \{\epsilon_2, \epsilon_1, a_y, a_z, a_x, [\mathbf{r}_2, \mathbf{r}_3, \mathbf{r}_1], \mathbf{t}\}\\
    \boldsymbol{\theta}^c_2 = \{\epsilon_2, \epsilon_1, a_z, a_x, a_y, [\mathbf{r}_3, \mathbf{r}_1, \mathbf{r}_2], \mathbf{t}\}
\end{aligned}
\end{equation}
It can be easily verified that both $[\mathbf{r}_2, \mathbf{r}_3, \mathbf{r}_1]$ and $[\mathbf{r}_3, \mathbf{r}_1, \mathbf{r}_2]$ are proper rotation matrices (orthogonal with determinant equals to 1).
When $\epsilon_1 = \epsilon_2$, the shapes encoded by the candidate parameters are identical to the current estimation.
This can be proved by substituting the candidate parameters into the implicit function of superquadrics.
In more general cases, $\boldsymbol{\theta}^c_1$ and $\boldsymbol{\theta}^c_2$ provides two superquadrics similar to $\boldsymbol{\theta}$, and thus hold similar likelihoods when evaluated by the probabilistic model.
Examples are visualized in Fig. \ref{fig:axis_mismatch}.

\textbf{Duality Similarity}:
\begin{figure*}[!htp]
    \centering 
    \includegraphics[scale=0.06]{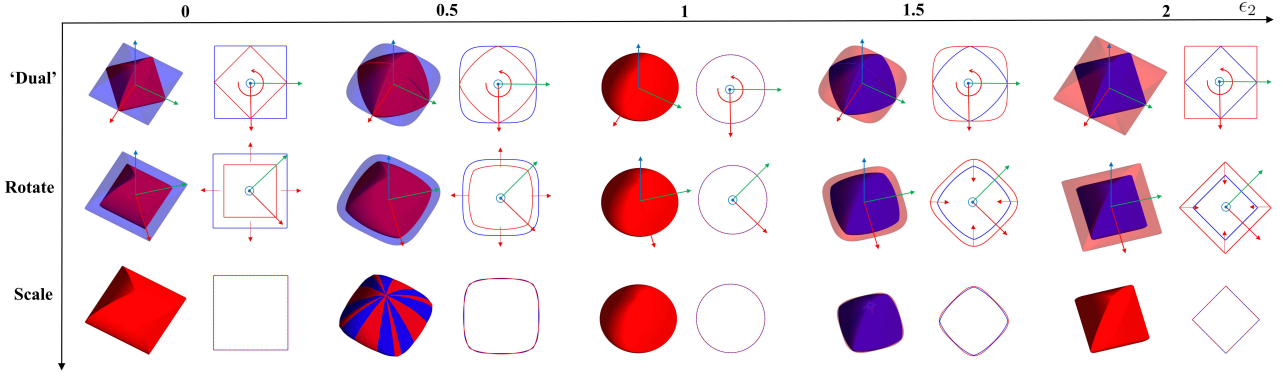} 
    \caption{Generate candidate superquadrics with duality similarity. The blue superquadrics are the current estimation. The red superquadrics demonstrate how similar superquadrics are generated by a sequence of transformations.}
    \label{fig:duality}
\vspace{-0.4cm}
\end{figure*}
The duality similarity is more complex compared with the axis-mismatch similarity.
To well illustrate the idea, we first review a property of superquadrics. 
When viewed from the superquadric frame (that is, not considering the spatial pose of the superquadric), a superquadric can be interpreted as the spherical product of two superellipses.
This property can be shown by expressing the superquadric surface with its parametric equation
\begin{equation}
\mathbf{p}(\eta, \omega)=
\left[\begin{array}{c}
a_{x} \cos^{\epsilon_{1}}{\eta} \cos^{\epsilon_{2}}{\omega} \\
a_{y} \cos^{\epsilon_{1}}{\eta} \sin^{\epsilon_{2}}{\omega} \\
a_{z} \sin^{\epsilon_{1}}{\eta}
\end{array}\right]
\label{eq:parametric}
\end{equation}
where $\mathbf{p}(\eta, \omega)$ is a bijective function which maps a point on a unit sphere $[\cos\eta, \sin\omega] \in \mathbb{S}^2$ to a point on the superquadric surface (parameterized by $\{\epsilon_1, \epsilon_2, a_x, a_y, a_z\}$).
It can be observed that
\begin{equation}
\mathbf{p}(\eta, \omega)=
\left[\begin{array}{c}
\cos^{\epsilon_{1}}{\eta}\\
a_{z} \sin^{\epsilon_{1}}{\eta}
\end{array}\right] \otimes
\left[\begin{array}{c}
a_{x} \cos^{\epsilon_{2}}{\omega} \\
a_{y} \sin^{\epsilon_{2}}{\omega}
\end{array}\right]
\label{eq:spherical_product}
\end{equation}
where $\otimes$ denotes the spherical product.
The right-hand side of the equation indicates a spherical product of two superellipses (paramaterized by $\{\epsilon_1, 1, a_z\}$ and $\{\epsilon_2, a_x, a_y\}$, respectively).
We call the former one the superellipse along the principal axis (z-axis), and the latter one the superellipse orthogonal to the principal axis.

Now that we have decoupled a superquadric into a pair of superellipses, it is natural to ask: can we find a similar superquadric by exploiting similar superellipses?
Actually, this goal can be achieved by utilizing the parametric ambiguity of the superellipse orthogonal to the principal axis ($\{\epsilon_2, a_x, a_y\}$).
When $a_x = a_y = \Bar{a}$, we can always obtain a similar or even identical superellipse by
(1) generating a `dual' superellipse $\{2-\epsilon_2, \Bar{a}, \Bar{a}\}$;
(2) rotating $\pi/4$ about the centroid;
and (3) scaling $\Bar{a}$ with a scaler $s$.
Then, we are able to obtain a similar superquadric by conducting the spherical product
\begin{equation*}
\resizebox{.95\hsize}{!}{$
\left[\begin{array}{c}
\cos^{\epsilon_{1}}{\eta}\\
a_{z} \sin^{\epsilon_{1}}{\eta}
\end{array}\right] \otimes
\left(
\left[\begin{array}{c c}
\cos\frac{\pi}{4} & -\sin\frac{\pi}{4}\\
\sin\frac{\pi}{4} & \cos\frac{\pi}{4}
\end{array}\right]
\left[\begin{array}{c}
s\cdot\Bar{a} \cos^{2-\epsilon_{2}}{\omega} \\
s\cdot\Bar{a} \sin^{2-\epsilon_{2}}{\omega}
\end{array}\right]\right)$}
\end{equation*}
Examples are shown in Fig. \ref{fig:duality}.
When $\epsilon_2=0$, $\epsilon_2=1$ or $\epsilon_2=2$, the generated superquadric is identical to the original one (Eq. \eqref{eq:spherical_product}).
In more general cases, the generated superquadric provides a candidate similar to the original one.
By taking the pose of the superquadric into consideration and relaxing $a_x=a_y$ with $a_x\approx a_y$ (in this paper, we assume $a_x\approx a_y$ when $0.8<|a_x/a_y|<1.2$), we can obtain a general formulation of the duality similarity as follows
\begin{equation}
\boldsymbol{\theta}^c_3= \{\epsilon_1, 2 - \epsilon_2, s\cdot \Bar{a}, s\cdot \Bar{a}, a_z, \mathbf{R}\cdot\mathbf{R}_z(\pi/4), \mathbf{t}\}
\end{equation}
where
\begin{equation*}
\begin{aligned}
&s= 
\left\{
\begin{array}{lcl}
    \left(\left(1 - \sqrt{2}\right)\epsilon_2 + \sqrt{2}\right) & if\; \epsilon_2 \leq 1\\
    \left(\sqrt{2}/2 - 1\right)\epsilon_2 + 2 - \sqrt{2}/2 & otherwise
\end{array}
\right.\\
&\Bar{a} = (a_x+a_y)/2
\end{aligned}
\end{equation*}
$s$ is the scale which compensates the expansion (when $\epsilon_2>1$) or shrinkage (when $\epsilon_2<1$) induced by the duality transformation ($2-\epsilon_2\rightarrow\epsilon_2$).
$\mathbf{R}_z(\pi/4)$ denotes a rotation of $\pi/4$ about the z-axis.

\textbf{Combinations of Similarities}:
Similar superquadrics can also be obtained by combining the axis-mismatch similarity with the duality similarity.
That is: 

(1) when $a_y\approx a_z$, we re-assign the principal axis to the y-axis and then look for its duality similarity
\begin{equation}
    \boldsymbol{\theta}^c_4= \{\epsilon_2, 2 - \epsilon_1, s\cdot \Bar{a}, s\cdot \Bar{a}, a_x, [\mathbf{r}_2, \mathbf{r}_3, \mathbf{r}_1]\cdot\mathbf{R}_z(\pi/4), \mathbf{t}\}
\end{equation}
where
\begin{equation*}
\begin{aligned}
&s= 
\left\{
\begin{array}{lcl}
    \left(\left(1 - \sqrt{2}\right)\epsilon_1 + \sqrt{2}\right) & if\; \epsilon_1 \leq 1\\
    \left(\sqrt{2}/2 - 1\right)\epsilon_1 + 2 - \sqrt{2}/2 & otherwise
\end{array}
\right.\\
&\Bar{a} = (a_y+a_z)/2
\end{aligned}
\end{equation*}

(2) similarly, when $a_x\approx a_z$, we re-assign the principal axis to the y-axis and then look for its duality similarity
\begin{equation}
    \boldsymbol{\theta}^c_5= \{\epsilon_2, 2 - \epsilon_1, s\cdot \Bar{a}, s\cdot \Bar{a}, a_y, [\mathbf{r}_3, \mathbf{r}_1, \mathbf{r}_2]\cdot\mathbf{R}_z(\pi/4), \mathbf{t}\}
\end{equation}
where
\begin{equation*}
\begin{aligned}
&s= 
\left\{
\begin{array}{lcl}
    \left(\left(1 - \sqrt{2}\right)\epsilon_1 + \sqrt{2}\right) & if\; \epsilon_1 \leq 1\\
    \left(\sqrt{2}/2 - 1\right)\epsilon_1 + 2 - \sqrt{2}/2 & otherwise
\end{array}
\right.\\
&\Bar{a} = (a_x+a_z)/2
\end{aligned}
\end{equation*}

\textbf{Switching Strategy}:
Utilizing the similarities, `highways' are built within the parameter space, connecting distant parameters encoding superquadrics with similar geometric shapes.
The S-step is triggered when the relative decrease of the negative log-likelihood is less than a threshold $\delta$.
Utilizing the similarities, a set of candidate parameters $\{\boldsymbol{\theta}_i^c\}$ are generated based on the current estimation $\boldsymbol{\theta}$.
Then, we check if the negative log-likelihood can be further decreased from a candidate.
If a valid candidate is found, we switch to it and continue the optimization;
Otherwise, we declare the current estimation $\boldsymbol{\theta}$ as optimal.
The detailed process of the S-step is summarized in Algorithm \ref{alg:s-step}.
\begin{algorithm}
\caption{S-step: geometric local optimum avoidance}\label{alg:s-step}
\begin{algorithmic}
    \State \textbf{Input:} $\boldsymbol{\theta}, \sigma^2, \mathbf{X}$ \Comment{current estimation and point cloud}
    \State \textbf{Output:} $\boldsymbol{\theta}_s, \sigma_s^2, success$ \Comment{switched parameter}
    \State $\boldsymbol{\theta}_s\leftarrow \boldsymbol{\theta}; \, \sigma^2_s\leftarrow\sigma^2$
    \State $success\leftarrow 0$
    \State $l\leftarrow Likelihood(\boldsymbol{\theta}, \sigma^2, \mathbf{X})$ \Comment{negative log-likelihood}
    \State $\{\boldsymbol{\theta}^c_i\}\leftarrow Similarities(\boldsymbol{\theta})$ \Comment{generate candidates}
    \For{$i=1,...,|\{\boldsymbol{\theta}^c_i\}|$}
        \State $\hat{\boldsymbol{\theta}}^c_i, \hat{\sigma}_i^2\leftarrow EM(\boldsymbol{\theta}^c_i, \sigma^2, \mathbf{X})$ \Comment{update candidate}
        \State $l_i\leftarrow Likelihood(\hat{\boldsymbol{\theta}}^c_i, \hat{\sigma}_i^2, \mathbf{X})$
        \If{$l_i<l$}
            \State $\boldsymbol{\theta}_s\leftarrow \hat{\boldsymbol{\theta}}^c_i;\, \sigma^2_s\leftarrow \hat{\sigma}^2_i$
            \State $success\leftarrow 1$
            \State $\textbf{Break}$
        \EndIf
    \EndFor
\end{algorithmic}
\end{algorithm}

\begin{figure} [!htp]
    \centering
    \includegraphics[width=0.85\columnwidth]{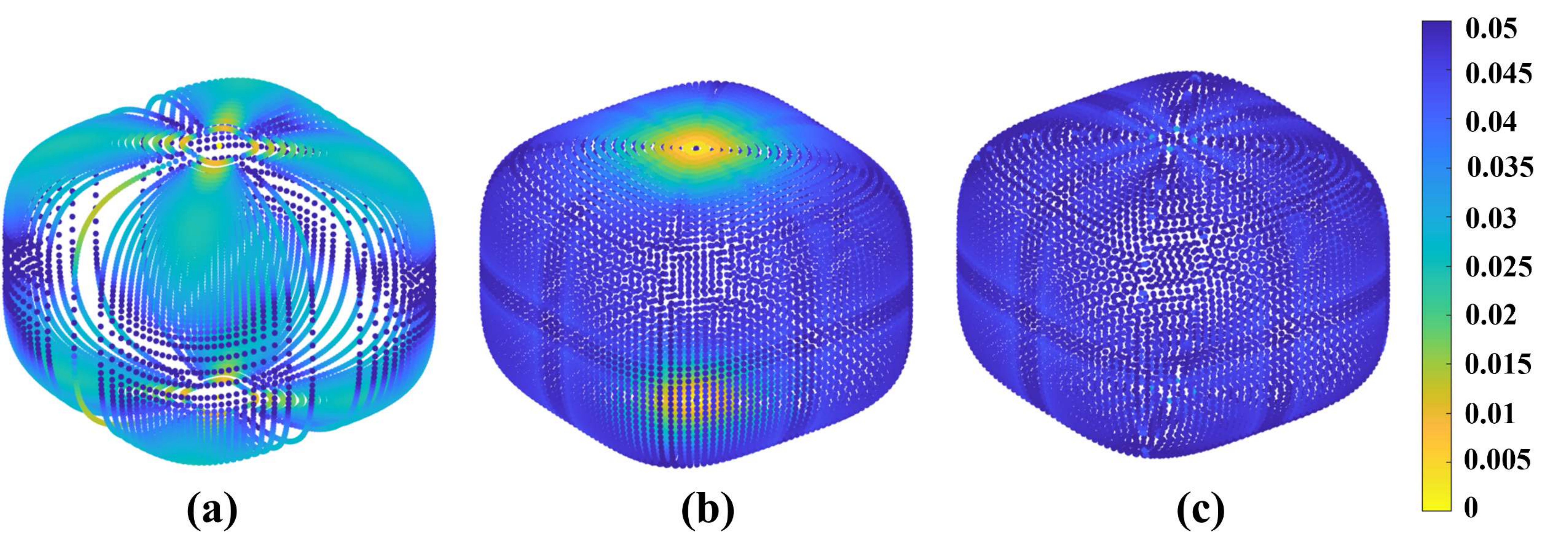} 
    \caption{Examples of different sampling strategies on a superquadric with shape parameters $\epsilon_1=0.5$ and $\epsilon_2=0.5$. The target sampling interval is $\Delta=0.05$.
    We use a heatmap to demonstrate the uniformity of the sampling results. The value of each point denotes the average Euclidean distances between the point and its four closest neighbors.
    (a-c) Results of the vanilla mapping, the method proposed by \cite{superquadrics_num_stable} and ours, respectively.}
    \label{fig:sampling}
    \vspace{-0.5cm}
\end{figure}

\section{Equal-distance Sampling on Superquadric Surface}
\label{sampling}
Eq. \eqref{eq:parametric} defines a bijective mapping between the unit sphere and a superquadric surface.
The most straightforward way of sampling points on a superquadric surface is to sample $\eta$ and $\omega$ within their range and then map to the target surface utilizing Eq. \eqref{eq:parametric}.
However, due to the non-linearity of the mapping, points evenly distributed in the spherical coordinates result in uneven samples on the superquadric surface.
In \cite{superquadrics_num_stable}, the authors propose a method to obtain a more homogeneous coverage of the surface.
As shown in Eq. \eqref{eq:spherical_product}, a superquadric is the spherical product of two superellipses.
Taking advantage of this property, two sets of 2D points are sampled independently on the periphery of each superellipse with a fixed interval\cite{pilu1995equal}.
Then, samples on the superquadric surface are constructed by taking the spherical product of the points between the sets.
As shown in Fig. \ref{fig:sampling}, the samples distribute more evenly compared with the vanilla mapping approach. 
However, the density gets significantly higher when the samples approach the poles along the z-axis.
As $\eta \rightarrow \frac{\pi}{2}/\frac{-\pi}{2}$, $|\cos^{\epsilon_1} \eta|\rightarrow 0$ and the perimeter of the sliced superellipse at higher latitude shrinks accordingly.
However, the superellipses at different latitudes contain the same number of points, and thus resulting in unevenly distributed samples after the spherical product.

To solve this problem, we improve the sampling strategy as follows.
Suppose the superquadric is parameterized by $\{\epsilon_1, \epsilon_2, a_x, a_y, a_z\}$, which can be decoupled into two superellipses $\{\epsilon_1, 1, a_z\}$ and $\{\epsilon_2, a_x, a_y\}$.
First, we sample the superellipse $\{\epsilon_1, 1, a_z\}$ with a fixed interval $\Delta$ using the algorithm in \cite{pilu1995equal}, and record the corresponding latitudes $\eta_i\in[-\pi/2, \pi/2], (i=1,2,...)$.
Then, instead of sampling directly on the superellipse $\{\epsilon_2, a_x, a_y\}$, we sample points evenly on re-scaled superellipses parameterized by $\{\epsilon_2, a_x\cos^{\epsilon_1}\eta_i, a_y\sin^{\epsilon_1}\eta_i\}$. 
In other words, we adaptively adjust the scale of the superellipse $\{\epsilon_2, a_x, a_y\}$ according to the current sampling latitude $\eta_i$.
And finally, by conducting the spherical product at each latitude, we obtain an overall equal-distance sampling of the superquadric surface.

\section{Implementation Details}
\label{experiment}
All the baseline methods are implemented with the official MATLAB Optimization Toolbox\cite{MatlabOTB}.
The step-tolerance (TolX) and optimality-tolerance (TolFun) are all set to $10^{-6}$.
Due to the numerical instability of Implicit-LSQ\cite{solina1990recovery}, Radial-LSQ\cite{GrossICCV88} and Robust-fitting\cite{hu1995robust}, the lower bounds of the shape parameters $\epsilon_1$ and $\epsilon_2$ are set to 0.1, as recommended by the authors.
Following the settings in\cite{superquadrics_num_stable}, the lower bounds of $\epsilon_1$ and $\epsilon_2$ are set to 0 for the NS method.

For the proposed method, the lower bounds of $\epsilon_1$ and $\epsilon_2$ are set to 0.
In Sec. 4.1 of the paper, the prior outlier probability $\omega_o=0$ for the partial data experiments, $\omega_o=0.2$ for the outlier experiments and $\omega_o=0.01$ for the noise experiments.
$\omega_o$ is set relatively low for the partial data and noise experiments, since no artificial outliers are added.
In Sec. 4.2, $\omega_o=0.01$ for the KIT dataset and $\omega_o=0.05$ for the BigBIRD dataset.
We set a higher outlier probability for the latter, since the point clouds are captured by a RGB-D camera, resulting in more outliers and noise.
$\omega_o$ is set to 0.8 in the Sec. 4.3, since we need to identify a large number of points as outliers in order to capture the major superquadric-like shape from a complex object.
Also in Sec. 4.3 of the paper, the maximum layer of the multi-superquadrics recovery is 3, and the cluster-pruning threshold is set to 60 points.
In all the experiments, the switching threshold $\delta=0.1$.


{\small
\bibliographystyle{ieee_fullname}
\bibliography{supp}
}